\documentclass{article}

\usepackage{microtype}
\usepackage{graphicx}
\usepackage{booktabs} 
\usepackage{multirow}

\usepackage{hyperref}

\usepackage{tikz}
\usepackage{pgfplots}
\pgfplotsset{compat=1.18}
\usepgfplotslibrary{fillbetween}
\usepackage{subcaption}

\usepackage[accepted]{icml2024}

\usepackage{units}
\usepackage{amsmath}
\usepackage{amssymb}
\usepackage{mathtools}
\usepackage{amsthm}
\usepackage{stmaryrd}

\usepackage[capitalize,noabbrev]{cleveref}

\theoremstyle{plain}
\newtheorem{theorem}{Theorem}[section]
\newtheorem{proposition}[theorem]{Proposition}

\theoremstyle{definition}

\theoremstyle{remark}
\newtheorem{remark}[theorem]{Remark}

\usepackage[textsize=tiny]{todonotes}

\newcommand*{\VEC}[1]  {\ensuremath{\boldsymbol{#1}}}
\newcommand*{\MAT}[1]  {\ensuremath{\boldsymbol{#1}}}

\newcommand*{\nfeatures}{\ensuremath{p}}
\newcommand*{\nsamples}{\ensuremath{n}}
\newcommand*{\nc}{\ensuremath{c}}
\newcommand*{\nmatrices}{\ensuremath{K}}
\newcommand*{\nclasses}{\ensuremath{Z}}

\newcommand*{\realSpace}{\ensuremath{\mathbb{R}}}
\newcommand*{\SymSpace}{\ensuremath{\mathcal{S}_{\nfeatures}}}
\newcommand*{\DiagSpace}{\ensuremath{\mathcal{D}_{\nfeatures}}}
\newcommand*{\SPDman}{\ensuremath{\mathcal{S}^{++}_{\nfeatures}}}
\newcommand*{\DPDman}{\ensuremath{\mathcal{D}^{++}_{\nfeatures}}}

\newcommand*{\eye}{\ensuremath{\MAT{I}_{\nfeatures}}}
\newcommand*{\onevec}{\ensuremath{\VEC{1}_{\nfeatures}}}

\newcommand*{\Cov}{\ensuremath{\MAT{C}}}
\newcommand*{\SCM}{\ensuremath{\MAT{\hat{C}}}}
\newcommand*{\linearLW}{\ensuremath{\MAT{\hat{C}}_{\textup{LW}}}}
\newcommand*{\nonlinearLW}{\ensuremath{\MAT{\hat{C}}_{\textup{LW-NL}}}}
\newcommand*{\CovRMTdist}{\ensuremath{\MAT{\hat{C}}_{\textup{dist}}}}

\newcommand*{\Mean}{\ensuremath{\MAT{G}}}
\newcommand*{\MeanRMT}{\ensuremath{\MAT{\hat{G}}_{\textup{RMT}}}}
\newcommand*{\MeanSCM}{\ensuremath{\MAT{\hat{G}}_{\textup{SCM}}}}
\newcommand*{\MeanLinearLW}{\ensuremath{\MAT{\hat{G}}_{\textup{LW}}}}
\newcommand*{\MeanNonLinearLW}{\ensuremath{\MAT{\hat{G}}_{\textup{LW-NL}}}}

\newcommand*{\point}{\ensuremath{\MAT{R}}}

\newcommand*{\eigsMat}{\ensuremath{\MAT{\Lambda}}}

\newcommand*{\tangentVector}{\ensuremath{\boldsymbol{\xi}}}
\newcommand*{\tangentVectorBis}{\ensuremath{\boldsymbol{\eta}}}

\newcommand*{\RMTdistSCMdeter}{\ensuremath{\hat{\delta}^2}}
\newcommand*{\RMTdistSCMs}{\ensuremath{\tilde{\delta}^2}}

\DeclareMathOperator{\tr}{tr}
\DeclareMathOperator{\diff}{d}

\DeclareMathOperator{\diag}{diag}
\DeclareMathOperator*{\argmin}{argmin}
\DeclareMathOperator*{\argmax}{argmax}
\DeclareMathOperator{\expm}{expm}
\DeclareMathOperator{\logm}{logm}

\newcommand*{\grad}{\ensuremath{\nabla}}

\newcommand*{\dataMat}{\ensuremath{\MAT{X}}}

\icmltitlerunning{Random matrix theory improved Frechet mean of symmetric positive definite matrices}

\begin{document}

\definecolor{darkgray}{RGB}{75,75,75}
\definecolor{myblue}{HTML}{1f77b4}
\definecolor{myorange}{HTML}{ff7f0e}
\definecolor{myred}{HTML}{d62728}

\twocolumn[
\icmltitle{Random matrix theory improved Fréchet mean of symmetric positive definite matrices}



\icmlsetsymbol{equal}{*}

\begin{icmlauthorlist}
\icmlauthor{Florent Bouchard}{l2s}
\icmlauthor{Ammar Mian}{listic}
\icmlauthor{Malik Tiomoko}{huawei}
\icmlauthor{Guillaume Ginolhac}{listic}
\icmlauthor{Frédéric Pascal}{l2s}
\end{icmlauthorlist}

\icmlaffiliation{huawei}{Huawei Paris Research Center}
\icmlaffiliation{l2s}{Université Paris Saclay, CNRS, CentraleSupélec, L2S}
\icmlaffiliation{listic}{Université Savoie Mont Blanc, LISTIC}

\icmlcorrespondingauthor{Florent Bouchard}{florent.bouchard@cnrs.fr}

\icmlkeywords{Random matrix theory, covariance matrices, Fréchet mean, Riemannian geometry}

\vskip 0.3in
]



\printAffiliationsAndNotice{\icmlEqualContribution} 

\begin{abstract}
In this study, we consider the realm of covariance matrices in machine learning, particularly focusing on computing Fréchet means on the manifold of symmetric positive definite matrices, commonly referred to as Karcher or geometric means.
Such means are leveraged in numerous machine learning tasks.
Relying on advanced statistical tools, we introduce a random matrix theory based method that estimates Fréchet means, which is particularly beneficial when dealing with low sample support and a high number of matrices to average.
Our experimental evaluation, involving both synthetic and real-world EEG and hyperspectral datasets, shows that we largely outperform state-of-the-art methods.
\end{abstract}

\section{Introduction}

Covariance matrices are of significant interest in machine learning, especially in scenarios with a limited number of labeled data or when dealing with high intra-class variability, as seen in EEG \cite{barachant2011multiclass} and remote sensing \cite{breizhcrops2020}. Numerous machine learning algorithms have been developed when features are covariance matrices, and therefore symmetrical positive-definite matrices (SPD).
A common and notable algorithm in this realm is the well-established nearest centroid. SPD matrices  find their use in deep learning networks \cite{huang2017riemannian,brooks2019riemannian}, metric learning \cite{zadeh2016,pmlr-v70-harandi17a}, domain adaptation \cite{kobler2022spd}, privacy protection \cite{reimherr2021}.
A pivotal component in most machine learning algorithms that utilize SPD matrices is the computation of a class barycenter. For SPD matrices, this barycenter is known as the Fréchet mean (or Karcher mean) \cite{bhatia}.
This mean is used, for example, for nearest centroid \cite{tuzel2008pedestrian}, pooling in SPD deep learning networks \cite{brooks2019riemannian} and metric learning \cite{zadeh2016}.
The optimal solution is not available analytically necessitating the use of iterative algorithms often based on deriving a Riemannian gradient \cite{boumal2023introduction}.
These algorithms are grounded in  Riemannian geometry, since matrices belong to specific manifolds depending on their specific properties (fair SPD, low rank, \emph{etc.}) and the chosen metric.
The geometry is often the classical one given for SPD matrices, but alternatives geometries are available to perform this algorithm such as Bures-Wassertein \cite{NEURIPS2021_4b04b0dc}, log-Euclidean \cite{utpala2023} and even for a more general manifold \cite{lou2020}.


These algorithms generally perform effectively, yet there are instances where the solution may be numerically unfeasible, particularly with the presence of a singular matrix. In this case, the most common solution is to regularize each of the covariance matrices. There is a plethora of work in this field.  The most common regularization technique involves shrinking the covariance estimate towards the identity matrix, introducing a parameter upon which the new estimate hinges. Numerous methods have been proposed to optimally estimate this parameter according to a chosen criterion.  A seminal contribution in this domain is by \cite{ledoit2004well}, where the mean square error (MSE) between the true covariance and the regularized covariance is used. The optimal parameter is finally calculated on the basis of statistical consistency considerations. Improvements have been proposed in \cite{ledoit2015spectrum} and \cite{ledoit2020analytical}. Extensions to non-Gaussian data have also been proposed \cite{ollila2014regularized,pascal2014regularized}.


In \cite{tiomoko2019random}, a novel approach was introduced, utilizing a distance-based criterion. This method draws upon the innovative distance presented in \cite{couillet2019random}, which offers a consistent estimation of the true distance between two matrices. This new estimate is derived from the tools of random matrix theory, which enables us to study the statistical behavior of the eigenvalues and eigenvectors of random matrices in a high-dimensional regime (when the size of the data $p$ and the number of samples $n$ grow at the same rate). While this new estimator demonstrates promising potential in terms of estimation, it is not without its practical challenges, such as the selection of initial values and the definition of an appropriate stopping criterion. A critical issue is its non-compliance with certain conditions set forth in \cite{tiomoko2019random} concerning the independence of the two matrices in the distance. Furthermore, similar to traditional regularization methods, this approach only regularizes the eigenvalues, leaving the eigenspaces unaltered. As indicated by empirical results, this form of regularization alone may not suffice to deliver optimal performance in classification or clustering tasks.


We recognize the significance of tailoring regularization strategies to specific applications. For example, in \cite{Kammoun2017}, the criterion is not the MSE or a distance, but maximizing the probability of detection. Given the primary goal of detection, this tailored approach yields substantially better outcomes compared to conventional regularization techniques. So, based on the results of \cite{couillet2019random,tiomoko2019random}, we propose a new regularization strategy directly related to classification algorithms.
In particular, we are interested in barycenters, which are for instance central to nearest centroid and K-means algorithms.
Specifically, our contributions are the followings:
\begin{itemize}
    \item First, we improve the RMT based covariance estimator proposed in~\cite{tiomoko2019random}. 
    \item Second, the main contribution is to propose a new RMT corrected Fréchet mean of SPD matrices by exploiting the improved distance from~\cite{couillet2019random}.
    \item Third, we adapt learning algorithms, \emph{i.e.}, K-means and nearest centroid classifier, to our new RMT mean.
    \item Finally, the interest of our approach is proved on both simulated and real EEG and hyperspectral data.
\end{itemize}


To ensure reproducibility, the code for the experiments discussed is accessible at \url{https://github.com/AmmarMian/icml-rmt-2024}.

\section{Preliminaries}

\subsection{Random matrix theory}
Random matrix theory (RMT) is a tremendous tool when it comes to studying the statistical behaviour of random matrices when the number of features $\nfeatures$ and the number of samples $\nsamples$ grow at the same rate toward infinity, \emph{i.e.}, as $\nfeatures,\nsamples\to\infty$, $\nfeatures/\nsamples\to\nc>0$.
In particular, from the seminal works~\cite{wishart1928generalised, marchenko1967distribution, silverstein1995empirical}, we know that the eigenvectors and eigenvalues of the sample covariance matrix (SCM) are not consistent in the large dimensional regime.
This lead researchers to regularize the SCM, more specifically its eigenvalues, in order to obtain consistent estimators; see~\emph{e.g.}~\cite{ledoit2015spectrum, ledoit2018optimal}.
Recently, with the rise of machine learning, distances between covariance matrices have attracted attention; see~\emph{e.g.}~\cite{couillet2019random, couillet2022random}. In the same spirit as for the study of covariance matrices, it has been shown that the distances between SPD matrices are not consistent and that it is then possible to regularize them to obtain improved distances which are then consistent in the high-dimensional regime.

\subsubsection{Covariance estimation}

Let $\dataMat\in\realSpace^{\nfeatures\times\nsamples}$ with true covariance $\Cov$ and SCM $\SCM=\frac1{\nsamples}\dataMat\dataMat^T$.
The most famous -- and probably the simplest -- way to regularize the SCM $\SCM$ consists in the linear shrinkage: $\linearLW = \rho\eye + \sqrt{1-\rho^2}\SCM$~\cite{ledoit2004well}.
Parameter $\rho>0$ is chosen so that it minimizes the expected $\ell2$ distance $\mathbb{E}[\| \Cov - \linearLW \|_2]$ asymptotically.
To estimate $\rho$ consistently, basic results from RMT are used.
However, in this setting, the eigenvalues are then biased. Another solution is to obtain a consistent estimate of the true eigenvalues $\lambda_i(\Cov)$.
A method to estimate these eigenvalues was first proposed in~\cite{KAR08}, with little success as the optimization process was very unstable.
This was solved in~\cite{ledoit2015spectrum} and~\cite{ledoit2018optimal}, where the $\ell2$ distance and a Stein loss are leveraged to estimate the $\lambda_i(\Cov)$'s from the $\lambda_i(\SCM)$'s with the so-called QuEST method.
The major limitation is that, even though QuEST is quite accurate, it is computationnally very expensive, which makes it complicated to employ in real scenarios.
Recently, \cite{ledoit2020analytical} proposed an analytical non-linear shrinkage of the $\lambda_i(\SCM)$'s, \emph{i.e.}, functions $\phi_i$ are learnt such that the $\lambda_i(\Cov)$'s are estimated through $\phi_i(\lambda_i(\SCM))$.
To determine the $\phi_i$'s, RMT, oracle non-linear shrinkage function and kernels are exploited.
They are chosen to minimize the true variance.
This method features the accuracy of QuEST while being numerically very efficient.

\subsubsection{Distance estimation}
\label{sec:preliminaries:RMT_dist}
Covariance matrices are increasingly exploited in machine learning algorithms such as Quadratic Discriminant Analysis, Metric Learning, Nearest Centroid, \emph{etc.}
Often, in such scenarios, covariance matrices are mainly leveraged to compute some kind of distance.
In this paper, we focus on distances between covariance matrices.
Unfortunately, as shown in~\cite{couillet2019random}, these are not consistent in the large dimensionality regime.
Some efforts have thus been dedicated to finding some good estimators.

In~\cite{couillet2019random,pereira2023consistent}, RMT corrected estimators of the squared distance between the true covariance matrices of some data are derived.
They considered two different cases.
In the first one, random data $\dataMat_1$ and $\dataMat_2$ with true covariance $\Cov_1$ and $\Cov_2$ and SCMs $\SCM_1$ and $\SCM_2$ are considered.
A consistent estimator $\RMTdistSCMs(\SCM_1,\SCM_2)$ of the squared distance $\delta^2(\Cov_1,\Cov_2)$ are derived.
In the other case, only one matrix is random.
We have $\dataMat$ with covariance $\Cov$ and SCM $\SCM$ and a deterministic SPD matrix $\point$.
A consistent estimator $\RMTdistSCMdeter(\point,\SCM)$ of the squared distance $\delta^2(\point,\Cov)$ is provided.
In both cases, estimators for a wide range of distances are obtained in closed form.
The main limitation of these RMT squared distance estimators is that they are valid only when data $\dataMat_1$ and $\dataMat_2$ are independent (respectively that $\point$ is not constructed with $\dataMat$).

We focus on the squared Fisher distance~\cite{skovgaard1984riemannian}, which is, for all $\Cov_1$ and $\Cov_2\in\SPDman$ (manifold of SPD matrices),
\begin{equation}
    \begin{array}{rcl}
        \delta^2(\Cov_1,\Cov_2) & = & \frac1{2\nfeatures}\| \logm(\Cov_1^{-1}\Cov_2) \|_2^2  \\
         & = & \frac1{2\nfeatures}\sum_{i=1}^{\nfeatures} \log^2(\lambda_i(\Cov_1^{-1}\Cov_2)),
    \end{array}
\label{eq:Fisher_dist}
\end{equation}
where $\logm(\cdot)$ denotes the matrix logarithm.
In the present work, we exploit the estimator $\RMTdistSCMdeter$ between some random matrix and a deterministic matrix in $\SPDman$ derived in~\cite{couillet2019random}.
It is provided in Theorem~\ref{thm:rmt_dist}.
\begin{theorem}[RMT corrected squared Fisher distance from~\cite{couillet2019random}]
    Given $\dataMat\in\realSpace^{\nfeatures\times\nsamples}$ ($\nfeatures>\nsamples$) with SCM $\SCM$ and a deterministic $\point\in\SPDman$, the RMT correction of the squared Fisher distance~\eqref{eq:Fisher_dist} is
    \begin{multline}
        \RMTdistSCMdeter(\point,\SCM) = \frac1{2\nfeatures}\sum_{i=1}^{\nfeatures} \log^2(\lambda_i)
        + \frac1{\nfeatures}\sum_{i=1}^{\nfeatures} \log(\lambda_i)
        \\
        - (\VEC{\lambda}-\VEC{\zeta})^T[ \frac1{\nfeatures}\MAT{Q}\onevec +\frac{1-\nc}{\nc}\VEC{q} ]
        - \frac{1-\nc}{2\nc}\log^2(1-\nc),
    \label{eq:RMT_Fisher_SCM_deterministic}
    \end{multline}
    where $\nc=\nicefrac{\nfeatures}{\nsamples}<1$;
    $\VEC{\lambda}$ and $\VEC{\zeta}$ contain the eigenvalues of $\point^{-1}\SCM$ and $\eigsMat-\frac{\sqrt{\VEC{\lambda}}\sqrt{\VEC{\lambda}}^T}{\nsamples}$, with $\eigsMat=\diag(\VEC{\lambda})$;
    $\VEC{q}\in\realSpace^{\nfeatures}$ such that $q_i=\frac{\log(\lambda_i)}{\lambda_i}$;
    and $\MAT{Q}\in\realSpace^{\nfeatures\times\nfeatures}$ is the matrix such that
    \begin{equation*}
        Q_{ij} = \left\{ 
        \begin{matrix}
            \frac{\lambda_i\log\left(\frac{\lambda_i}{\lambda_j}\right) - (\lambda_i-\lambda_j)}{(\lambda_i-\lambda_j)^2}, & i\neq j
            \\
            \frac1{2\lambda_i}, & i=j
        \end{matrix}
        \right. .
    \end{equation*}
\label{thm:rmt_dist}
\end{theorem}

\subsection{Riemannian optimization on $\SPDman$}
\label{sec:preliminaries:Ropt}
Riemannian optimization~\cite{absil2009optimization,boumal2023introduction} provide generic methods to solve constrained optimization problems over any smooth manifold.
In the present work, we are interested in optimization on the manifold of SPD matrices $\SPDman$ and we are limiting ourselves to the Riemannian gradient descent algorithm.
Let $f:\SPDman\to\realSpace$ be an objective function.
The goal is to solve the optimization problem
\begin{equation*}
    \argmin_{\point\in\SPDman} \quad f(\point).
\end{equation*}
To do so, the differential structure of $\SPDman$ is exploited.
Since $\SPDman$ is open in the space of symmetric matrices $\SymSpace$, the tangent space at any point $\point\in\SPDman$ can be identified to $\SymSpace$.
The next step is to equip $\SPDman$ with a Riemannian metric.
The choice that appears natural in our case is the Fisher information metric of the normal distribution, which yields~\eqref{eq:Fisher_dist}. 
It is, for all $\point\in\SPDman$, $\tangentVector$, $\tangentVectorBis\in\SymSpace$,
\begin{equation}
    \langle \tangentVector, \tangentVectorBis \rangle_{\point} = 
    \tr(\point^{-1}\tangentVector\point^{-1}\tangentVectorBis).
\label{eq:Fisher_metric}
\end{equation}
It allows to define the Riemannian gradient $\grad f(\point)$ of $f$ at $\point\in\SPDman$ as the only matrix in $\SymSpace$ such that, for all $\tangentVector\in\SymSpace$,
\begin{equation}
    \langle \grad f(\point), \tangentVector \rangle_{\point} = \diff f(\point)[\tangentVector],
\end{equation}
where $\diff f(\point)[\tangentVector]$ denotes the directional derivative of $f$ at $\point$ in the direction $\tangentVector$.
The Riemannian gradient provides a descent direction of the cost function $f$ in the tangent space at $\point$.
From there, we need to obtain a new point on $\SPDman$.
This is achieved by a retraction $R$, which maps every tangent vector at any point onto the manifold.
In our opinion, the optimal retraction choice on $\SPDman$ is the second-order approximation of the Riemannian exponential mapping (generalization of a straight line on a manifold) defined in~\cite{jeuris2012survey}, for all $\point\in\SPDman$ and $\tangentVector\in\SymSpace$, as
\begin{equation}
    R_{\point}(\tangentVector) = \point + \tangentVector + \frac12\tangentVector\point^{-1}\tangentVector.
\label{eq:retr}
\end{equation}
All the tools to apply the Riemannian gradient descent algorithm in order to optimize the cost function $f$ have now been introduced.
Given an initial guess $\point_0\in\SPDman$, the sequence of iterates $\{\point_{\ell}\}$ produced by the gradient descent is given through the recurrence
\begin{equation}
    \point_{\ell+1} = R_{\point_\ell}(-t_\ell\grad f(\point_\ell)),
\label{eq:RgradDescent}
\end{equation}
where $t_\ell>0$ is the stepsize, which can be computed through a linesearch; see \emph{e.g.},~\cite{absil2009optimization}.

\section{Closely related work}

In this paper, we explore covariance and Fréchet mean estimation by leveraging~\eqref{eq:RMT_Fisher_SCM_deterministic}.
In~\cite{tiomoko2019random}, the problem of estimating covariance by exploiting~\eqref{eq:RMT_Fisher_SCM_deterministic} has already been considered.
Indeed, authors are interested in the optimization problem
\begin{equation}
    \argmin_{\point\in\SPDman} \quad \RMTdistSCMdeter(\point,\SCM),
\label{eq:RMT_cov_optim_problem}
\end{equation}
and also consider Riemannian optimization to solve it.
As previously explained, for~\eqref{eq:RMT_Fisher_SCM_deterministic} to provide an accurate approximation of $\delta^2(\point,\Cov)$, $\point$ must be sufficiently independent from $\dataMat$.
When trying to solve~\eqref{eq:RMT_cov_optim_problem}, this is a big issue.
The gradient obviously depends on $\dataMat$.
It inevitably induces some dependency between $\point$ and $\dataMat$ along iterations.
As a consequence, $\RMTdistSCMdeter(\point,\SCM)$ becomes irrelevant  at some point and lead to an inappropriate solution.

Concerning covariance estimation, the big difference between their approach and ours lies in how this issue is handled.
In~\cite{tiomoko2019random}, since $\RMTdistSCMdeter(\point,\SCM)$ becomes negative when it is no longer informative, they considered optimizing $\point\mapsto(\RMTdistSCMdeter(\point,\SCM))^2$.
As it did not appear to be sufficient, they also limited their search to the eigenvalues of the true covariance matrix, \emph{i.e.}, they assumed $\point=\MAT{U}\MAT{\Delta}\MAT{U}^T$, where $\MAT{U}$ contain the eigenvectors of the sample covariance $\SCM$ and $\MAT{\Delta}$ contain the sought eigenvalues.

In our work, we employ a very different strategy.
Indeed, we choose to keep $\point\mapsto\RMTdistSCMdeter(\point,\SCM)$ as the actual cost function and our search space remains $\SPDman$.
Instead of changing these, we wisely define a new stopping criterion.
Further notice that we derive the gradient in a very different way and end up with a formula that has a different form.
Finally, they do not consider at all the Fréchet mean estimation problem, which is clearly the main contribution of our paper.

\section{RMT improved covariance estimation}
\label{sec:cov}

In this section, we improve the RMT based covariance estimation proposed in \cite{tiomoko2019random}. This latter is based on the squared distance estimator~\eqref{eq:RMT_Fisher_SCM_deterministic}. The main stake of this section is to be able to disrupt as much as possible the dependency on $\dataMat$ that is created along the optimization process which leads to mitigate results in \cite{tiomoko2019random}.
To do so, in Section~\ref{sec:cov:algo}, we clean up the method proposed in~\cite{tiomoko2019random}, which relies on~\eqref{eq:RMT_Fisher_SCM_deterministic}.
More specifically, we don't take the square of the squared distance estimator, perform optimization on $\SPDman$, and propose a properly adapted stopping criterion.
%
Finally, in Section~\ref{sec:cov:simu}, simulations are performed in order to compare the proposed approach to baseline methods and concluding remarks are provided.

%

\subsection{Covariance estimator algorithm}
\label{sec:cov:algo}
We first consider the optimization problem~\eqref{eq:RMT_cov_optim_problem}, which leverages the squared distance estimator~\eqref{eq:RMT_Fisher_SCM_deterministic}.
In this scenario, the covariance estimator $\CovRMTdist$ is obtained by minimizing $f:\point\mapsto\RMTdistSCMdeter(\point,\SCM)$, which approximates $\point\mapsto\delta^2(\point,\Cov)$, where $\Cov$ is the true covariance of $\dataMat$.
To solve the optimization problem, we resort to Riemannian optimization on $\SPDman$ with the tools presented in Section~\ref{sec:preliminaries:Ropt}.
To be able to implement the Riemannian gradient descent, all we need is the Riemannian gradient of $f:\point\mapsto\RMTdistSCMdeter(\point,\SCM)$.

The objective $f$ is a function of the eigenvalues of $\point^{-1}\SCM$, \emph{i.e.}, $f(\point)=g(\eigsMat)$, where $\eigsMat\in\DPDman$  (manifold of positive definite diagonal matrices) contain the eigenvalues of $\point^{-1}\SCM$ (or equivalently $\point^{\nicefrac{-1}{2}}\SCM\point^{\nicefrac{-1}{2}}$ to keep a symmetric matrix).
First, in Proposition~\ref{prop:Rgrad_eigsfun}, the Riemannian gradient $\grad f(\point)$ of $f$ in $\SPDman$ is given as a function of the Riemannian gradient $\grad g(\eigsMat)$ of $g$ in $\DPDman$, also equipped with metric~\eqref{eq:Fisher_metric}.
As for $\SPDman$, $\grad g(\eigsMat)$ is the only element of the space of diagonal matrices $\DiagSpace$ such that, for all $\tangentVector\in\DiagSpace$,
    $\diff g(\eigsMat)[\tangentVector] = \tr(\eigsMat^{-1}\grad g(\eigsMat)\eigsMat^{-1}\tangentVector).$
%
\begin{proposition}
\label{prop:Rgrad_eigsfun}
    Let $\SCM\in\SPDman$ and $f:\SPDman\to\realSpace$ such that for all $\point\in\SPDman$, $f(\point)=g(\eigsMat)$, where $g:\DPDman\to\realSpace$ and $\eigsMat$ is obtained through the eigenvalue decomposition $\point^{\nicefrac{-1}{2}}\SCM\point^{\nicefrac{-1}{2}}=\MAT{U}\eigsMat\MAT{U}^T$.
    It follows that
    \begin{equation*}
        \grad f(\point) = -\point^{\nicefrac12}\MAT{U}\eigsMat^{-1}\grad g(\eigsMat)\MAT{U}^T\point^{\nicefrac12},
    \end{equation*}
    where $\grad g(\eigsMat)$ is the Riemannian gradient of $g$ at $\eigsMat$ in $\DPDman$.
\end{proposition}
\begin{proof}
    See Appendix~\ref{app:proofs}.
\end{proof}
It now remains to compute the Riemannian gradient $\grad g(\eigsMat)$ of $g$ at $\eigsMat\in\DPDman$, where $g$ corresponds to the RMT corrected squared Fisher distance~\eqref{eq:RMT_Fisher_SCM_deterministic}.
It is provided in Proposition~\ref{prop:RMT_Fisher_SCM_deterministic_grad}.
\begin{proposition}
\label{prop:RMT_Fisher_SCM_deterministic_grad}
    Let $g:\DPDman\to\realSpace$ the function such that $g(\eigsMat)=\RMTdistSCMdeter(\point,\SCM)$, with $\RMTdistSCMdeter$ defined in~\eqref{eq:RMT_Fisher_SCM_deterministic} and $\eigsMat$ the eigenvalues of $\point^{\nicefrac{-1}{2}}\SCM\point^{\nicefrac{-1}{2}}$.
    It follows that
    \begin{multline*}
        \hspace*{-8pt}
        \grad g(\eigsMat) = \frac1{\nfeatures}[\logm(\eigsMat)+\eye]\eigsMat
        - \eigsMat^2[\MAT{\Delta} + \diag(\MAT{A}\MAT{V}\MAT{\Delta}\MAT{V}^T)]
        \\
        - \frac1{\nfeatures}\eigsMat^2\diag(\MAT{B}\onevec(\VEC{\lambda}-\VEC{\zeta})^T + \onevec(\VEC{\lambda}-\VEC{\zeta})^T\MAT{C})
        \\
        -\frac{1-\nc}{\nc}(\eye-\logm(\eigsMat))(\eigsMat-\diag(\VEC{\zeta})),
    \end{multline*}
    where $\VEC{\zeta}$ and $\MAT{V}$ are the eigenvalues and eigenvectors of $\eigsMat-\frac{\sqrt{\VEC{\lambda}}\sqrt{\VEC{\lambda}}^T}{\nsamples}$; $\MAT{\Delta}=\diag(\frac1{\nfeatures}\MAT{Q}\onevec+\frac{(1-\nc)}{\nc}\VEC{q})$, with $\MAT{Q}$ and $\VEC{q}$ defined in~\eqref{eq:RMT_Fisher_SCM_deterministic}; and $\MAT{A}$, $\MAT{B}$ and $\MAT{C}$ are the matrices such that
    \begin{equation*}
        \MAT{A}_{ij} = \left\{ 
        \begin{matrix}
            -\frac1{\nsamples}\sqrt{\frac{\lambda_j}{\lambda_i}}, & i\neq j
            \\
            1-\frac1{\nsamples}, & i=j
        \end{matrix}
        \right.,
    \end{equation*}
    \begin{equation*}
        \MAT{B}_{ij} = \left\{ 
        \begin{matrix}
            -\frac{(\lambda_i+\lambda_j)\log(\frac{\lambda_i}{\lambda_j})}{(\lambda_i-\lambda_j)^3} - \frac{2}{(\lambda_i-\lambda_j)^2}, & i\neq j
            \\
            -\frac1{\lambda_i^2}, & i=j
        \end{matrix}
        \right.,
    \end{equation*}
    \begin{equation*}
        \MAT{C}_{ij} = \left\{ 
        \begin{matrix}
            \frac{1}{\lambda_j(\lambda_i-\lambda_j)} + \frac{2\lambda_i\log(\frac{\lambda_i}{\lambda_j})}{(\lambda_i-\lambda_j)^3} - \frac{2}{(\lambda_i-\lambda_j)^2}, & i\neq j
            \\
            \frac1{2\lambda_i^2}, & i=j
        \end{matrix}
        \right..
    \end{equation*}
\end{proposition}
\begin{proof}
    See Appendix~\ref{app:proofs}.
\end{proof}
Injecting Proposition~\ref{prop:RMT_Fisher_SCM_deterministic_grad} in Proposition~\ref{prop:Rgrad_eigsfun} yields the Riemannian gradient of $f$.
This is all that is needed to perform the Riemannian gradient descent~\eqref{eq:RgradDescent} on $\SPDman$ in order to solve~\eqref{eq:RMT_cov_optim_problem}.

For covariance estimation, the interest of solving~\eqref{eq:RMT_cov_optim_problem} lies in the fact that $\RMTdistSCMdeter(\point,\SCM)$ provides an accurate estimation of $\delta^2(\point,\Cov)$.
Unfortunately, $\RMTdistSCMdeter(\point,\SCM)$ does not actually approximate $\delta^2(\point,\Cov)$ for any $\point\in\SPDman$.
Indeed, if $\point$ is too related to $\dataMat$ (\emph{e.g.}, $\point=\SCM$), then $\RMTdistSCMdeter(\point,\SCM)$ is no longer informative.
In fact, it can even take negative values.
To handle this, \cite{tiomoko2019random} chose to rather perform optimization on the square of the RMT squared distance estimator~\eqref{eq:RMT_Fisher_SCM_deterministic}.
In this paper, we argue that this is not necessary and that wisely choosing the stopping criterion is enough.
Indeed, starting from an adequate initialization (\emph{i.e.}, one that is sufficiently independent from $\dataMat$), our idea is to pursue optimization while $\RMTdistSCMdeter(\point,\SCM)$ is relevant and to stop as we reach the limit.
From a statistical point of view, when $\point$ is not too related to $\dataMat$, one expects $\RMTdistSCMdeter(\point,\SCM)\geq O(\nicefrac{-1}{\nfeatures})$.
Thus, our new stopping criterion consists in checking that we have $f(\point)=\RMTdistSCMdeter(\point,\SCM)\geq\nicefrac{-\alpha}{\nfeatures}$, and to stop as soon as this is no longer true.
Some cross-validation on synthetic data for various $\nfeatures$ and $\nsamples$ lead us to believe that choosing $\alpha=10$ is the best option.
The method to estimate covariance by leveraging the RMT corrected squared distance is presented in Algorithm~\ref{algo:RMTCov}%
\footnote{
    Notice that the linesearch~\cite{absil2009optimization,boumal2023introduction} is slightly modified.
    In addition to the Armijo condition, we add the condition $f(R_{\point_{\ell}}(-t_{\ell}\grad f(\point_{\ell}))\geq\nicefrac{-\alpha}{\nfeatures}$ to the backtracking procedure on $t_{\ell}$.
}.

\begin{remark}
    Concerning initialization, we need to select one that is sufficiently unrelated to $\dataMat$.
    The simplest choice is $\eye$.
    The SCM $\SCM$ is of course not an option.
    The non-linear shrinkage estimator $\nonlinearLW$ from~\cite{ledoit2020analytical} also usually does not work.
    Interestingly, the linear shrinkage estimator $\linearLW$~\cite{ledoit2004well} appears to usually be the strongest option we considered.
\end{remark}

\begin{algorithm}[t]
    \caption{Covariance based on RMT corrected distance}
    \label{algo:RMTCov}
    \begin{algorithmic}
        \STATE {\bfseries Input:}
            data $\dataMat\in\realSpace^{\nfeatures\times\nsamples}$,
            initial guess $\point_0\in\SPDman$,
            tolerances $\alpha>0$, $\varepsilon>0$,
            maximum iterations $\ell_{\textup{max}}$
        \STATE Compute SCM $\SCM=\frac1{\nsamples}\dataMat\dataMat^T$
        \STATE Set $\ell=0$
        \REPEAT
        \STATE Compute gradient $\grad f(\point_{\ell})$ (Prop.~\ref{prop:Rgrad_eigsfun} and~\ref{prop:RMT_Fisher_SCM_deterministic_grad})
        \STATE Compute stepsize $t_{\ell}$ with linesearch
        \STATE $\point_{\ell+1} = R_{\point_{\ell}}(-t_{\ell}\grad f(\point_{\ell}))$, with $R$ defined in~\eqref{eq:retr}
        \STATE $\ell=\ell+1$
        \UNTIL{
            $f(\point_{\ell})<\nicefrac{-\alpha}{\nfeatures}$ 
            {\bfseries or}
            $\delta^2(\point_{\ell},\point_{\ell-1})<\varepsilon$
            {\bfseries or}
            $\ell>\ell_{\textup{max}}$
        }
        \STATE {\bfseries Return:} $\CovRMTdist=\point_{\ell}$
    \end{algorithmic}
\end{algorithm}

\subsection{Simulations summary and concluding remarks}
\label{sec:cov:simu}
Detailed simulations on covariance estimation are provided in Appendix~\ref{app:simu_RMTCov}.
Due to space limitations, only a summary and some concluding remarks are provided here. 
In our simulations, we randomly generate a covariance matrix.
We then simulate some data that are used to estimate their covariance.
Various methods are considered:
the SCM $\SCM$,
the linear Ledoit-Wolf estimator $\linearLW$~\cite{ledoit2004well},
the non-linear Ledoit-Wolf estimator $\nonlinearLW$~\cite{ledoit2020analytical},
and our RMT distance based method $\CovRMTdist$ from Algorithm~\ref{algo:RMTCov}.

The best performance is obtained with $\nonlinearLW$.
Our estimator $\CovRMTdist$ improves upon $\SCM$ and $\linearLW$ at low sample support.
Considering that it is also more expensive (others are analytical), it does not seem advantageous and exploiting~\eqref{eq:RMT_Fisher_SCM_deterministic} might not be suited for covariance estimation.
Notice however that in some rare cases at low sample support, $\SCM$, $\linearLW$ and $\nonlinearLW$ behave poorly while our estimator performs well.
We believe that this occurs when the SCM does not provide good eigenvectors.

\section{RMT corrected Fréchet mean on $\SPDman$}
\label{sec:mean}
This section contains the most interesting contribution of this paper.
We propose an original RMT based method to estimate the Fréchet mean (also known as Karcher or geometric mean) $\Mean\in\SPDman$ of a set of $\nmatrices$ covariance matrices $\{\Cov_k\}$ in $\SPDman$ when only some data $\{\dataMat_k\}$ in $\realSpace^{\nfeatures\times\nsamples}$ are known.
Notice that this corresponds to the setting that is always encountered in practice when one aims to exploit one or several Fréchet means of some covariance matrices in order to perform a learning task.
Usually, getting a Fréchet mean is achieved with a two steps procedure: (\emph{i}) covariance matrices are estimated from the data and (\emph{ii}) their mean is computed with an iterative method such as~\cite{fletcher2004principal,jeuris2012survey}.
The obtained Fréchet means are then exploited for classification or clustering, for instance in Nearest Centroid or K-Means algorithms; see \emph{e.g.},~\cite{tuzel2008pedestrian,barachant2011multiclass}.

In this work, we rather develop a one step method that directly estimate the mean $\Mean$ from observations $\{\dataMat_k\}$ without trying to obtain their covariance matrices.
As for our attempt on improving covariance in Section~\ref{sec:cov}, our model heavily relies on the RMT corrected squared Fisher distance~\eqref{eq:RMT_Fisher_SCM_deterministic}.
In Section~\ref{sec:mean:algo}, the optimization problem that we consider along with the algorithm proposed to solve it are presented.
In Section~\ref{sec:mean:learning}, our RMT mean is leveraged to define original Nearest Centroid and K-Means.
Finally, in Section~\ref{sec:mean:simu}, our method is compared with the usual two steps procedure for various covariance estimators on simulated data.
Concluding remarks are also provided.

\subsection{RMT mean algorithm}
\label{sec:mean:algo}
Let a set of $\nmatrices$ raw data matrices $\{\dataMat_k\}$ in $\realSpace^{\nfeatures\times\nsamples}$ with SCMs $\{\SCM_k\}$.
To obtain our RMT based Fréchet $\MeanRMT$ on $\SPDman$, we simply replace the squared Fisher distance $\delta^2$ defined in~\eqref{eq:Fisher_dist} with its RMT corrected counterpart~\eqref{eq:RMT_Fisher_SCM_deterministic} in the definition of the Fréchet mean.
It follows that $\MeanRMT$ is solution to the optimization problem
\begin{equation}
    \argmin_{\point\in\SPDman} \quad h(\point) = \frac1{\nmatrices}\sum_{k=1}^{\nmatrices} \RMTdistSCMdeter(\point,\SCM_k).
\label{eq:RMT_mean_optim_problem}
\end{equation}
The objective function $h:\point\mapsto\frac1{\nmatrices}\sum_k\RMTdistSCMdeter(\point,\SCM_k)$ aims to approximate the cost function one would get if the true covariance matrices $\{\Cov_k\}$ were known, \emph{i.e.}, $\point\mapsto\frac1{\nmatrices}\sum_k\delta^2(\point,\Cov_k)$.
Hence, our hope is to significantly improve the estimation of the true mean $\Mean$ as compared to two steps procedures that compute the Fréchet mean of some covariance estimators.
\begin{remark}
    When $\nsamples$ is large enough, the usual cost function $\point\mapsto\frac1{\nmatrices}\sum_k\delta^2(\point,\SCM_k)$ well approximates $\point\mapsto\frac1{\nmatrices}\sum_k\delta^2(\point,\Cov_k)$ since the SCM $\SCM_k$ asymptotically converges to the true covariance $\Cov_k$.
    It is no longer true for a small $\nsamples$.
    In comparison, our proposed cost function appears advantageous for a wider range of number of samples $\nsamples$.
\end{remark}

As for covariance estimation from Section~\ref{sec:cov}, it is crucial to determine whether our cost function is truly informative.
Given $k$, recall that $\RMTdistSCMdeter(\point,\SCM)_k$ well approximates $\delta^2(\point,\Cov_k)$ only if $\point$ is sufficiently independent from $\dataMat_k$.
Again, while optimizing $h$, some dependency on $\dataMat_k$ is introduced.
However, this time, the dependency on $\dataMat_k$ is counterbalanced by the ones on the other data matrices $\{\dataMat_{k'}\}_{k'\neq k}$.
Since data matrices are independent from one another, overall, we expect $\point$ to remain sufficiently independent from each $\dataMat_k$ as soon as $\nmatrices$ is large enough%
\footnote{
    In practice, it appears true even for small values of $\nmatrices$.
    Indeed, in our simulations (Section~\ref{sec:mean:simu}), even for $K=2$, we improve upon the SCM associated with the usual Fréchet mean on $\SPDman$.
}.

To solve~\eqref{eq:RMT_mean_optim_problem}, we again resort to a Riemannian gradient descent on $\SPDman$.
It is thus needed to compute the gradient of $h$.
Writing $h:\point\mapsto\frac1{\nmatrices}\sum_k f_k(\point)$, with $f_k:\point\mapsto\RMTdistSCMdeter(\point,\SCM_k)$, one has
\begin{equation}
    \grad h(\point) = \frac1{\nmatrices} \sum_{k=1}^K \grad f_k(\point),
\label{eq:RMT_mean_grad}
\end{equation}
where $\grad f_k(\point)$ is obtained by combining Propositions~\ref{prop:Rgrad_eigsfun} and~\ref{prop:RMT_Fisher_SCM_deterministic_grad}.
With the tools of Section~\ref{sec:preliminaries:Ropt}, it is enough to implement the Riemannian gradient descent.
Our proposed method is summarized in Algorithm~\ref{algo:RMTMean}.

\begin{remark}
    The complexity of an iteration of Algorithm~\ref{algo:RMTMean} is of the same order of magnitude as an iteration of the Riemannian gradient descent for the usual Fréchet mean on $\SPDman$ (\emph{i.e.}, with~\eqref{eq:Fisher_dist}).
    The difference between the two lies in gradients computations.
    Even though~\eqref{eq:RMT_mean_grad} appears way more complicated, it is not that much more expensive.
    Concerning costly operations, in both cases, we have to perform a Cholesky decomposition and its inverse (to compute $\point^{\nicefrac12}$ and $\point^{\nicefrac{-1}2}$), and $\nmatrices$ eigenvalue decompositions (of $\point^{\nicefrac{-1}2}\SCM_k\point^{\nicefrac{-1}2}$).
    To get~\eqref{eq:RMT_mean_grad}, we further need $\nmatrices$ eigenvalue decompositions (of $\eigsMat-\frac{\VEC{\lambda}\VEC{\lambda}^T}{n}$).
    The rest only involve less expensive operations (matrix multiplications, \emph{etc.}).
\end{remark}

\begin{algorithm}
    \caption{RMT corrected Fréchet mean on $\SPDman$}
    \label{algo:RMTMean}
    \begin{algorithmic}
        \STATE {\bfseries Input:}
            data $\{\dataMat_k\}_{k=1}^{\nmatrices}$ in $\realSpace^{\nfeatures\times\nsamples}$,
            initial guess $\point_0\in\SPDman$,
            tolerance $\varepsilon>0$,
            maximum iterations $\ell_{\textup{max}}$
        \FOR{$k$ {\bfseries in} $\llbracket1,\nmatrices\rrbracket$}
        \STATE Compute SCM $\SCM_k=\frac1{\nsamples}\dataMat_k\dataMat_k^T$
        \ENDFOR
        \STATE Set $\ell=0$
        \REPEAT
        \STATE Compute gradient $\grad h(\point_{\ell})$ with~\eqref{eq:RMT_mean_grad}
        \STATE Compute stepsize $t_{\ell}$ with linesearch
        \STATE $\point_{\ell+1} = R_{\point_{\ell}}(-t_{\ell}\grad h(\point_{\ell}))$, with $R$ defined in~\eqref{eq:retr}
        \STATE $\ell=\ell+1$
        \UNTIL{
            $\delta^2(\point_{\ell},\point_{\ell-1})<\varepsilon$
            {\bfseries or}
            $\ell>\ell_{\textup{max}}$
        }
        \STATE {\bfseries Return:} $\MeanRMT=\point_{\ell}$
    \end{algorithmic}
\end{algorithm}

\subsection{Nearest Centroid and K-Means based on RMT}
\label{sec:mean:learning}
To exploit the RMT corrected Fréchet mean on $\SPDman$ in learning, we adapt the acclaimed Nearest Centroid classifying and K-Means clustering methods.
Both algorithms rely on the RMT Fréchet mean $\MeanRMT$ and on the corrected squared Fisher distance $\RMTdistSCMdeter$.

In the supervised Nearest Centroid setting, provided in Algorithm~\ref{algo:NearestCentroid}, we have a training set $\{\dataMat_k,y_k\}_{k=1}^{\nmatrices}$, where each $\dataMat_k\in\realSpace^{\nfeatures\times\nsamples}$ belongs to a class $y_k$ in $\llbracket 1,\nclasses \rrbracket$.
In the fitting phase, the RMT Fréchet means $\{\MeanRMT^{(z)}\}$ of every class $z\in\llbracket1,\nclasses\rrbracket$ are learnt by solving
\begin{equation}
    \MeanRMT^{(z)} = \argmin_{\point\in\SPDman} \quad \frac1{\nmatrices_z}\sum_{y_k\in\mathcal{A}_z} \RMTdistSCMdeter(\point,\SCM_k),
\end{equation}
where $\mathcal{A}_z=\{y_k: k\in\llbracket1,\nmatrices\rrbracket \textup{ and } y_k=z\}$ and $\nmatrices_z$ is the cardinal of $\mathcal{A}_z$.
They are obtained with Algorithm~\ref{algo:RMTMean}.
Then, in the prediction phase, given some unlabeled data $\dataMat\in\realSpace^{\nfeatures\times\nsamples}$ with SCM $\SCM$, the decision rule is
\begin{equation}
    y = \argmin_{z\in\llbracket1,\nclasses\rrbracket} \quad \{\RMTdistSCMdeter(\MeanRMT^{(z)},\SCM)\}_{z=1}^{\nclasses}.
\label{eq:NearestCentroid_prediction}
\end{equation}

\begin{algorithm}[t]
    \caption{Nearest Centroid classifier based on RMT}
    \label{algo:NearestCentroid}
    \begin{algorithmic}
        \STATE {\bfseries Fitting phase}
        \\ \vspace*{-5pt} \hrulefill 
        \STATE {\bfseries Input:}
            data $\{\dataMat_k\}_{k=1}^{\nmatrices}$ in $\realSpace^{\nfeatures\times\nsamples}$,
            labels $\{y_k\}_{k=1}^{\nmatrices}$ in $\llbracket1,\nclasses\rrbracket$
        \FOR{$z$ {\bfseries in} $\llbracket 1,\nclasses \rrbracket$}
        \STATE Compute $\MeanRMT^{(z)}$ from $\{\dataMat_k: y_k=z\}$ with Algo.~\ref{algo:RMTMean}
        \ENDFOR
        \STATE {\bfseries Return:} $\{\MeanRMT^{(z)}\}_{z=1}^{\nclasses}$
        \\ \hrulefill
        \STATE {\bfseries Prediction phase}
        \\ \vspace*{-5pt} \hrulefill
        \STATE {\bfseries Input:}
            unlabeled data $\dataMat\in\realSpace^{\nfeatures\times\nsamples}
            $
        \STATE Compute SCM $\SCM=\frac1{\nsamples}\dataMat\dataMat^T$
        \FOR{$z$ {\bfseries in} $\llbracket 1,\nclasses \rrbracket$}
        \STATE Compute $\RMTdistSCMdeter(\MeanRMT^{(z)},\SCM)$ with~\eqref{eq:RMT_Fisher_SCM_deterministic}
        \ENDFOR
        \STATE Compute $y$ with~\eqref{eq:NearestCentroid_prediction} 
        \STATE {\bfseries Return:} label $y\in\llbracket1,\nclasses\rrbracket$
    \end{algorithmic}
\end{algorithm}

The Nearest Centroid classifier can be adapted to the unsupervised K-Means clustering scenario, detailed in Algorithm~\ref{algo:KMeans}.
In this setting, one has a set of $\nmatrices$ data samples $\{\dataMat_k\}$ in $\realSpace^{\nfeatures\times\nsamples}$.
Given a certain number of classes $\nclasses$, the goal is to assign a label $y_k\in\llbracket1,\nclasses\rrbracket$ to each $\dataMat_k$.
This is achieved iteratively.
Each iteration $\ell$ consists of two steps.
An assignment step, where, given $\nclasses$ means $\{\MeanRMT^{(z)}(\ell)\}$, a label $y_k(\ell)$ is assigned to each $\dataMat_k$ leveraging rule~\eqref{eq:NearestCentroid_prediction}.
A mean update step, where every $\MeanRMT^{(z)}(\ell)$ is recomputed from $\{\dataMat_k: y_k(\ell)=z\}$.
This is repeated until we reach some equilibrium.
It is well known that the results of this procedure are very sensitive to the initialization of centroids.
As prescribed in~\cite{arthur2007k} in the Euclidean case, we consider using several initializations and keep results from the one maximizing the criterion
\begin{equation}
    \mathcal{I}(\{\mathbf{X}_k, y_k\}, \{\MeanRMT^{(z)}\}) = \sum_{k=1}^{\nmatrices} \RMTdistSCMdeter(\MeanRMT^{(y_i)},\dataMat_k).
    \label{eq:KMeans_inertia}
\end{equation}

\begin{algorithm}[t]
    \caption{K-Means clustering based on RMT}
    \label{algo:KMeans}
    \begin{algorithmic}
        \STATE {\bfseries Input:}
            data $\{\dataMat_k\}_{k=1}^{\nmatrices}$ in $\realSpace^{\nfeatures\times\nsamples}$,
            number of classes $\nclasses$,
            tolerance $\alpha>0$,
            maximum iterations $\ell_{\textup{max}}$,
            number of different initializations $M$
        \FOR{$m$ {\bfseries in} $\llbracket1,M\rrbracket$}
        \STATE Randomly choose $\{k_z\}_{z=1}^{\nclasses}$ and set $\MeanRMT^{(z)}(0)=\SCM_{k_z}$
        \STATE Compute $\{y_k(0)\}$ with~\eqref{eq:NearestCentroid_prediction}
        \STATE Set $\ell=0$
        \REPEAT
            \STATE $\ell=\ell+1$
            \STATE Compute $\{\MeanRMT^{(z)}(\ell)\}$ from $\{\dataMat_k: y_k(\ell-1)=z\}$ with Algo.~\ref{algo:RMTMean}
            \STATE Compute $\{y_k(\ell)\}$ with~\eqref{eq:NearestCentroid_prediction} 
        \UNTIL{
            $\frac1{\nmatrices}\sum_k \| y_k(\ell) - y_k(\ell-1) \|_2 < \alpha$
            {\bfseries or}
            $\ell > \ell_{\textup{max}}$
        }
        \STATE Compute inertia $\mathcal{I}(m)$ for initialization $m$ with~\eqref{eq:KMeans_inertia}
        \ENDFOR
        \STATE Compute $m_{\textup{max}} = \argmax_{m\in\llbracket1,M\rrbracket} \,\, \{\mathcal{I}(m)\}_{m=1}^M$.
        \STATE {\bfseries Return:} $\{y_k\}_{k=1}^{\nmatrices}$ associated with  $m_{\textup{max}}$.
    \end{algorithmic}
\end{algorithm}

\subsection{Simulations}
\label{sec:mean:simu}

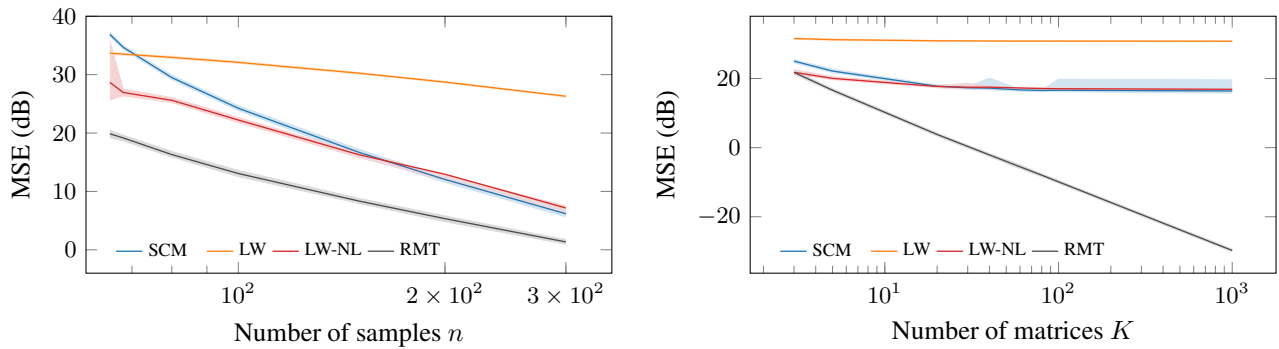
\begin{figure*}
    \begin{center}
\begin{tikzpicture}

    \begin{semilogxaxis}[
        width  =0.5\linewidth,
        height =5cm,
        at     ={(0,0)},
        xlabel={Number of samples $\nsamples$},
        xmin=60, xmax=350,
        minor xtick={70,80,90,200,300},
        xtick={100,200, 300},
        xticklabels={
          \(\textstyle {10^{2}}\),
          \(\textstyle {2\times 10^{2}}\),
          \(\textstyle {3\times 10^{2}}\),
        },
        ylabel={MSE~(dB)},
        ymin=-4, ymax=40,
        tick label style={font=\footnotesize},
        legend cell align={left},
        legend columns=4, 
        legend style={draw=none,fill=none, font=\scriptsize, at={(0.04, 0.01)},anchor=south west, /tikz/column 2/.style={column sep=5pt,}},
        legend image code/.code={\draw [#1] (0cm,0cm) -- (0.35cm,0cm);},
    ]
        \addplot[color=myblue!20,line width=0.1pt, forget plot, name path=A] table [x=n_samples,y expr = 20*log10(\thisrow{SCM}),col sep=comma] {./Figures/mse_nsamples/5.csv};
        \addplot[color=myblue!20,line width=0.1pt, forget plot, name path=B] table [x=n_samples,y expr = 20*log10(\thisrow{SCM}),col sep=comma] {./Figures/mse_nsamples/95.csv};
        \addplot[myblue!20, forget plot] fill between[of=A and B];

        \addplot[color=myorange!20,line width=0.1pt, forget plot, name path=A] table [x=n_samples,y expr = 20*log10(\thisrow{LW_linear}),col sep=comma] {./Figures/mse_nsamples/5.csv};
        \addplot[color=myorange!20,line width=0.1pt, forget plot, name path=B] table [x=n_samples,y expr = 20*log10(\thisrow{LW_linear}),col sep=comma] {./Figures/mse_nsamples/95.csv};
        \addplot[myorange!20, forget plot] fill between[of=A and B];

        \addplot[color=myred!20,line width=0.1pt, forget plot, name path=A] table [x=n_samples,y expr = 20*log10(\thisrow{LW_nonlinear}),col sep=comma] {./Figures/mse_nsamples/5.csv};
        \addplot[color=myred!20,line width=0.1pt, forget plot, name path=B] table [x=n_samples,y expr = 20*log10(\thisrow{LW_nonlinear}),col sep=comma] {./Figures/mse_nsamples/95.csv};
        \addplot[myred!20, forget plot] fill between[of=A and B];

        \addplot[color=darkgray!20,line width=0.1pt, forget plot, name path=A] table [x=n_samples,y expr = 20*log10(\thisrow{RMT}),col sep=comma] {./Figures/mse_nsamples/5.csv};
        \addplot[color=darkgray!20,line width=0.1pt, forget plot, name path=B] table [x=n_samples,y expr = 20*log10(\thisrow{RMT}),col sep=comma] {./Figures/mse_nsamples/95.csv};
        \addplot[darkgray!20, forget plot] fill between[of=A and B];
        
        \addplot[color=myblue,line width=0.5pt] table [x=n_samples,y expr = 20*log10(\thisrow{SCM}),col sep=comma] {./Figures/mse_nsamples/mean.csv};
        \addlegendentry{SCM};

        \addplot[color=myorange,line width=0.5pt] table [x=n_samples,y expr = 20*log10(\thisrow{LW_linear}),col sep=comma] {./Figures/mse_nsamples/mean.csv};
        \addlegendentry{LW};

        \addplot[color=myred,line width=0.5pt] table [x=n_samples,y expr = 20*log10(\thisrow{LW_nonlinear}),col sep=comma] {./Figures/mse_nsamples/mean.csv};
        \addlegendentry{LW-NL};

        \addplot[color=darkgray,line width=0.5pt] table [x=n_samples,y expr = 20*log10(\thisrow{RMT}),col sep=comma] {./Figures/mse_nsamples/mean.csv};
        \addlegendentry{RMT};
    \end{semilogxaxis}

    \begin{semilogxaxis}[
        width  =0.5\linewidth,
        height =5cm,
        at     ={(0.515\linewidth,0)},
        xlabel={Number of matrices $\nmatrices$},
        ylabel={MSE~(dB)},
        tick label style={font=\footnotesize},
        legend cell align={left},
        legend columns=4, 
        legend style={draw=none,fill=none, font=\scriptsize, at={(0.04, 0.01)},anchor=south west, /tikz/column 2/.style={column sep=5pt,}},
        legend image code/.code={\draw [#1] (0cm,0cm) -- (0.35cm,0cm);},
    ]

        \addplot[color=myblue!20,line width=0.1pt, forget plot, name path=A] table [x=n_matrices,y expr = 20*log10(\thisrow{SCM}),col sep=comma] {./Figures/mse_nmatrices/64_128/5.csv};
        \addplot[color=myblue!20,line width=0.1pt, forget plot, name path=B] table [x=n_matrices,y expr = 20*log10(\thisrow{SCM}),col sep=comma] {./Figures/mse_nmatrices/64_128/95.csv};
        \addplot[myblue!20, forget plot] fill between[of=A and B];

        \addplot[color=myorange!20,line width=0.1pt, forget plot, name path=A] table [x=n_matrices,y expr = 20*log10(\thisrow{LW_linear}),col sep=comma] {./Figures/mse_nmatrices/64_128/5.csv};
        \addplot[color=myorange!20,line width=0.1pt, forget plot, name path=B] table [x=n_matrices,y expr = 20*log10(\thisrow{LW_linear}),col sep=comma] {./Figures/mse_nmatrices/64_128/95.csv};
        \addplot[myorange!20, forget plot] fill between[of=A and B];

        \addplot[color=myred!20,line width=0.1pt, forget plot, name path=A] table [x=n_matrices,y expr = 20*log10(\thisrow{LW_nonlinear}),col sep=comma] {./Figures/mse_nmatrices/64_128/5.csv};
        \addplot[color=myred!20,line width=0.1pt, forget plot, name path=B] table [x=n_matrices,y expr = 20*log10(\thisrow{LW_nonlinear}),col sep=comma] {./Figures/mse_nmatrices/64_128/95.csv};
        \addplot[myred!20, forget plot] fill between[of=A and B];

        \addplot[color=darkgray!20,line width=0.1pt, forget plot, name path=A] table [x=n_matrices,y expr = 20*log10(\thisrow{RMT}),col sep=comma] {./Figures/mse_nmatrices/64_128/5.csv};
        \addplot[color=darkgray!20,line width=0.1pt, forget plot, name path=B] table [x=n_matrices,y expr = 20*log10(\thisrow{RMT}),col sep=comma] {./Figures/mse_nmatrices/64_128/95.csv};
        \addplot[darkgray!20, forget plot] fill between[of=A and B];
        
        \addplot[color=myblue,line width=0.5pt] table [x=n_matrices,y expr = 20*log10(\thisrow{SCM}),col sep=comma] {./Figures/mse_nmatrices/64_128/mean.csv};
        \addlegendentry{SCM};

        \addplot[color=myorange,line width=0.5pt] table [x=n_matrices,y expr = 20*log10(\thisrow{LW_linear}),col sep=comma] {./Figures/mse_nmatrices/64_128/mean.csv};
        \addlegendentry{LW};

        \addplot[color=myred,line width=0.5pt] table [x=n_matrices,y expr = 20*log10(\thisrow{LW_nonlinear}),col sep=comma] {./Figures/mse_nmatrices/64_128/mean.csv};
        \addlegendentry{LW-NL};

        \addplot[color=darkgray,line width=0.5pt] table [x=n_matrices,y expr = 20*log10(\thisrow{RMT}),col sep=comma] {./Figures/mse_nmatrices/64_128/mean.csv};
        \addlegendentry{RMT};
    
    \end{semilogxaxis}
\end{tikzpicture}
\vspace*{-15pt}
\end{center}
    \caption{Mean square error (MSE) over 1000 trials of the estimated Fréchet mean towards the true mean matrix with respect to the number of samples $\nsamples$ (left) and number of matrices $\nmatrices$ (right).
    Parameters are $\nfeatures=64$, $\nmatrices=10$ on the left and $\nsamples=128$ on the right.
    Lines correspond to the medians while filled areas correspond to the $5^{\textup{th}}$ and $95^{\textup{th}}$ quantiles.}
    \label{fig:mse_mean}
\end{figure*}

This section contains simulations conducted to evaluate the performance of our proposed RMT based method as compared to state-of-the-art algorithms.
The experimental setup is as follows.
A center $\Mean=\MAT{U}\MAT{\Delta}\MAT{U}^T\in\SPDman$ is generated, where $\MAT{U}$ is uniformly drawn on $\mathcal{O}_{\nfeatures}$ (orthogonal group), and $\MAT{\Delta}$ is randomly drawn on $\DPDman$.
Maximal and minimal diagonal entries of $\MAT{\Delta}$ are set to $\sqrt{a}$ and $\nicefrac{1}{\sqrt{a}}$, where $a=100$ is the condition number.
Remaining non-zero elements are uniformly drawn in-between.
Then, $\nmatrices$ matrices $\{\Cov_k\}$ whose Fréchet mean is $\Mean$ are randomly generated.
To do so, given $k$, $\frac{\nfeatures(\nfeatures+1)}{2}$ values are drawn from $\mathcal{N}(0,\sigma^2)$, with $\sigma^2=0.1$.
These are used to canonically construct $\MAT{S}_k\in\SymSpace$.
A set of $\nmatrices$ centered symmetric matrices $\{\tangentVector_k\}$ is obtained by canceling the mean of the $\MAT{S}_k$'s, \emph{i.e.}, $\tangentVector_k = \MAT{S}_k - \frac1{\nmatrices}\sum_{k'}\MAT{S}_{k'}$.
Hence, $\frac1{K}\sum_k\tangentVector_k=\MAT{0}$.
Finally, $\Cov_k=\Mean^{\nicefrac12}\expm(\tangentVector_k)\Mean^{\nicefrac12}$, where $\expm(\cdot)$ denotes the matrix exponential.
After that, we generate $\nmatrices$ matrices $\dataMat_k$ in $\realSpace^{\nfeatures\times\nsamples}$ such that each column of $\dataMat_k$ is drawn from $\mathcal{N}(\VEC{0},\Cov_k)$.

To estimate $\Mean$ from $\{\dataMat_k\}$, several methods are considered.
First, two steps methods are employed.
They consist in estimating covariance matrices and then their usual Fréchet mean.
The mean resulting from the SCM estimator is denoted $\MeanSCM$.
The ones obtained after employing the linear and non-linear Ledoit-Wolf estimators are denoted $\MeanLinearLW$ and $\MeanNonLinearLW$, respectively.
These are compared to our proposed RMT based mean $\MeanRMT$ obtained with Algorithm~\ref{algo:RMTMean}.
To measure performance, we use the squared Fisher distance~\eqref{eq:Fisher_dist} between the true mean and its estimator.

Results are presented in Figure~\ref{fig:mse_mean}.
These indicate a distinct advantage of our proposed RMT-based method over the others across all examined scenarios.
One can observe that when the number of samples $\nsamples$ grows, $\MeanSCM$ and $\MeanNonLinearLW$ slowly catch up with $\MeanRMT$.
When $\nsamples$ is fixed (moderately low) and the number of matrices $\nmatrices$ increases, $\MeanRMT$ is the only one that strongly improves.
In conclusion, our RMT-based method demonstrates superior performance, especially when the sample size $\nsamples$ is moderately limited and the number of matrices $\nmatrices$ is large.

\section{Real data learning experiments}

\begin{table*}[t]
    \centering
    \begin{tabular}{rcccc}
        & SCM & LW & LW-NL & RMT \\
        \cline{2-5}
        GrosseWentrup09 & 0.632 $\pm$ 0.0867& 0.624 $\pm$ 0.0829 & $\times$ & {\bf 0.638 $\pm$ 0.0917} \\
        Schirmeister17 & 0.597 $\pm$  0.139 & 0.483 $\pm$ 0.0958 & 0.561 $\pm$ 0.120 & {\bf 0.603 $\pm$ 0.120} \\
        Cho17 & 0.615 $\pm$ 0.158 & 0.609 $\pm$ 0.136 & 0.601 $\pm$ 0.131 & {\bf 0.622 $\pm$ 0.158} \\
        Lee19 & {\bf 0.666 $\pm$ 0.138} & 0.642 $\pm$ 0.130 & 0.626 $\pm$ 0.126 & 0.66 $\pm$ 0.137\\
    \end{tabular}
    \caption{Classification results on EEG motor imaging data.}
    \label{tab:eeg_mi}
\end{table*}

\begin{table*}[t]
\centering
\begin{tabular}{rcccccccc}
                              &               &                       & \multicolumn{2}{c}{SCM} & \multicolumn{2}{c}{LW} & \multicolumn{2}{c}{RMT}         \\
                              & $\nfeatures$ & $\nsamples$ & acc             & mIoU  & acc          & mIoU        & acc            & mIoU           \\ \cline{2-9} 
\multirow{4}{*}{Indian pines} & 5             & 5$\times$5            & 0.385           & 0.278 & 0.302        & 0.204       & \textbf{0.454} & \textbf{0.367} \\
                              & 16            & 5$\times$5            & 0.357           & 0.229 & 0.316        & 0.215       & \textbf{0.413} & \textbf{0.284} \\ & 24            & 7$\times$7            & 0.377           & 0.253 & 0.359        & 0.248       & \textbf{0.453} & \textbf{0.285} \\
                              \cline{2-9} 
\multirow{4}{*}{Salinas}      & 5             & 5$\times$5            & 0.542           & 0.382 & 0.402        & 0.252       & \textbf{0.777} & \textbf{0.631} \\
                              & 10            & 7$\times$7            & 0.525           & 0.34  & 0.449        & 0.303       & \textbf{0.746} & \textbf{0.532} \\
                              & 16            & 11$\times$11          & 0.497           & 0.317 & 0.404        & 0.244       & \textbf{0.632} & \textbf{0.461} \\ \cline{2-9} 
Pavia                         & 5             & 5$\times$5            & 0.629           & 0.378 & 0.615        & 0.319       & \textbf{0.819} & \textbf{0.549} \\ \cline{2-9} 
KSC                           & 5             & 5$\times$5   & 0.263           & 0.167 & 0.247        & 0.169       & \textbf{0.377} & \textbf{0.222}
\end{tabular}
\caption{Clustering results for hyperspectral data. For Indian pines, we did 10 initializations and 5 for the other datasets.}
\label{tab:hyperspectral}
\end{table*}

To assess the practical relevance of the proposed RMT-based mean estimation method, two real-world scenarios are considered:
(\emph{i}) electroencephalography (EEG) classification using Nearest Centroid classifiers and (\emph{ii}) clustering of hyperspectral images using K-Means algorithms.
Various strategies were implemented for mean computation and distance. Specifically, for mean estimation, we consider two step strategies where we first estimate covariances and then compute their generic Fréchet mean associated with~\eqref{eq:Fisher_dist}.
As before, we consider the SCM, linear Ledoit-Wolf (LW) and non-linear Ledoit-Wolf (LW-NL) estimators.
These methods were then benchmarked against our proposed RMT-based Nearest Centroid and K-Means algorithms, as detailed in Section~\ref{sec:mean:learning}. The development and evaluation of these methods were conducted in Python. Specifically, SCM and LW implementations were sourced from the scikit-learn library~\cite{pedregosa2018scikitlearn}, while LW-NL comes from scikit-RMT\footnote{\url{https://scikit-rmt.readthedocs.io/}}. The conventional Fréchet means, standard Nearest Centroid and K-Means algorithms were taken from the pyRiemann library~\cite{alexandre_barachant_2023_8059038}.

\subsection{EEG data}

We initiated our analysis by assessing the Nearest Centroid classifier's efficacy on EEG data, specifically focusing on motor imagery datasets accessible via the MOABB platform~\cite{Aristimunha_Mother_of_all_2023}.
In this context, subjects participate in experiments where they are instructed to mentally simulate various movements, encompassing actions like the motion of the left or right hand, feet, tongue, among others.
The following datasets are used:
GrosseWentrup2009, where $\nclasses=2$, $\nfeatures=128$, signals resampled to $100$Hz;
Schirmeister2017, where  $\nclasses=4$, $\nfeatures=128$, signals resampled to $100$Hz;
Cho2017, where  $\nclasses=2$, $\nfeatures=64$, signals resampled to $128$Hz, trials taken from 1s to 3s;
Lee2019, where  $\nclasses=2$, $\nfeatures=62$, signals resampled to $100$Hz, trials taken from 2s to 3s.


The outcomes are summarized in Table~\ref{tab:eeg_mi}. An analysis of the results reveals that the SCM and RMT methods demonstrate comparable levels of performance across all datasets, with RMT achieving marginal enhancements in three out of the four datasets. Conversely, the accuracy rates for both LW and LW-NL are notably lower. Specifically, in the case of GrosseWentrup2009, the LW-NL method encountered issues, failing to produce SPD matrices as required, rendering it non-functional. This observation underscores the superior reliability of the RMT method as a regularization technique for these datasets. However, given the minimal performance gap between RMT and SCM, the incremental benefit of RMT may not justify the additional complexity for this particular application.

\subsection{Hyperspectral data}

Our second experiment with real data delves into the clustering of hyperspectral remote sensing datasets, including Indian Pines, Salinas, Pavia, and KSC%
\footnote{
    Available at \url{https://www.ehu.eus/ccwintco/index.php/Hyperspectral_Remote_Sensing_Scenes}.
}.
 These datasets are inherently diverse, characterized by a unique number of bands and classes. They also feature annotated ground truths. Certain zones labeled as "undefined" are considered unreliable and hence are omitted from the accuracy calculations of the clustering methods. Nevertheless, these zones are included during the clustering phase to ensure realistic evaluation. 

 Data preprocessing involves three main steps: normalizing data by subtracting the image's global mean, employing Principal Component Analysis (PCA) to select a set number of channels $\nfeatures$ as per prior research~\cite{9627641}, and using a sliding window with overlap for data sampling around each pixel. We excluded the LW-NL method due to numerical instability. The K-means algorithm, capped at $100$ iterations with early stopping at a $10^{-4}$ tolerance, concludes with a linear assignment optimization to align the clustered image with ground truth, optimizing classification accuracy.

The results, detailed in Table~\ref{tab:hyperspectral}, evaluate classification accuracy and mean intersection over union. Our RMT method consistently outperforms SCM and LW across all datasets and varying feature/sample sizes. In our opinion, this success can be attributed to the high number of matrices per class in these datasets, resonating with our simulation insights: in such contexts, the RMT-corrected mean significantly enhances accuracy. In essence, for data scenarios with extensive matrices per class, the RMT approach proves highly effective.

\subsection{Discussion}
The first thing that we notice on these real data learning experiments is the is the lower performance of the methods associated with the linear and non-linear Ledoit-Wolf estimators.
For LW, this is in accordance with our simulations.
It is likely that this is because the true covariance matrices are in fact quite far from the identity matrix.
So, linearly shrinking toward the identity does not improve estimation in this case.
For LW-NL, this is more surprising.
On these real data, LW-NL has proved itself numerically unstable and unable to consistently provide SPD matrices.

Concerning our RMT-based method, we observe that:
\begin{itemize}
    \item When the number of samples $\nsamples$ is sufficiently big as compared to the dimension $\nfeatures$, our proposed RMT mean becomes equivalent to the usual Fisher mean.
    In this case, one can expect to obtain very similar results with our method as compared to the method associated with the SCM.
    This is indeed what we observe with EEG data, for which a rather large $\nsamples$ is available (and needed to capture the information contained in the data).
    \item When $\nsamples$ is comparable to $\nfeatures$, then our performance depends on the number of matrices $\nmatrices$ available to compute the mean:
    \begin{itemize}
        \item[$\circ$] when $\nmatrices$ is very small, the issue is that there is a clear statistical dependence introduced between the estimated mean and the data at hand.
        In this case, the RMT distance estimator can become irrelevant and our mean is not very satisfying.
        This is the worst case scenario for our method and its main limitation.
        However, in our simulations, a small number of matrices $\nmatrices$ is enough for our method to be competitive.
        \item[$\circ$] when $\nmatrices$ becomes big, this is where our estimator performs very well.
        In fact, on simulated data, our RMT estimator appears to converge to the true mean as $\nmatrices$ grows while other methods do not.
        For the real hyperspectral data that we consider, we are exactly in this scenario and the performance we obtain corresponds to our simulations, \emph{i.e.}, our method performs really well as compared to other methods.
    \end{itemize}
\end{itemize}

\section{Conclusions and perspectives}
\label{sec:conclusion}


The first part of this paper presents a refined regularized covariance estimator, building upon the corrected squared distance outlined in \cite{couillet2019random}. While this work aligns closely with \cite{tiomoko2019random}, it introduces subtle yet noteworthy enhancements, including a more comprehensive treatment of matrix independence and a new stopping criterion rooted in statistical principles. The primary contribution, however, is the development of a novel Fréchet mean algorithm tailored for random matrices under conditions of low sample support, utilizing a Riemannian gradient on the SPD matrix manifold. When applied to Nearest Centroid classifiers and K-means clustering, this new method demonstrates great potential. It appears very advantageous when dealing with a large number of matrices per class, offering a big improvement over traditional methods in this case.

\section*{Acknowledgement}
This research was supported by DATAIA Convergence Institute as part of the ``Programme d’Investissement d’Avenir", (ANR-17-CONV-0003) operated by Laboratoire des Signaux et Systèmes.

\section*{Impact Statement}
This paper presents work whose goal is to advance the field of Machine Learning. There are many potential societal consequences of our work, none which we feel must be specifically highlighted here.

\bibliographystyle{icml2024}
\bibliography{references.bib}

\begin{thebibliography}{38}
\providecommand{\natexlab}[1]{#1}
\providecommand{\url}[1]{\texttt{#1}}
\expandafter\ifx\csname urlstyle\endcsname\relax
  \providecommand{\doi}[1]{doi: #1}\else
  \providecommand{\doi}{doi: \begingroup \urlstyle{rm}\Url}\fi

\bibitem[Absil et~al.(2009)Absil, Mahony, and Sepulchre]{absil2009optimization}
Absil, P.-A., Mahony, R., and Sepulchre, R.
\newblock \emph{Optimization algorithms on matrix manifolds}.
\newblock Princeton University Press, 2009.

\bibitem[Aristimunha et~al.(2023)Aristimunha, Carrara, Guetschel, Sedlar,
  Rodrigues, Sosulski, Narayanan, Bjareholt, Quentin, Schirrmeister, Kalunga,
  Darmet, Gregoire, Abdul~Hussain, Gatti, Goncharenko, Thielen, Moreau, Roy,
  Jayaram, Barachant, and Chevallier]{Aristimunha_Mother_of_all_2023}
Aristimunha, B., Carrara, I., Guetschel, P., Sedlar, S., Rodrigues, P.,
  Sosulski, J., Narayanan, D., Bjareholt, E., Quentin, B., Schirrmeister,
  R.~T., Kalunga, E., Darmet, L., Gregoire, C., Abdul~Hussain, A., Gatti, R.,
  Goncharenko, V., Thielen, J., Moreau, T., Roy, Y., Jayaram, V., Barachant,
  A., and Chevallier, S.
\newblock {Mother of all BCI Benchmarks}, 2023.
\newblock URL \url{https://github.com/NeuroTechX/moabb}.

\bibitem[Arthur \& Vassilvitskii(2007)Arthur and Vassilvitskii]{arthur2007k}
Arthur, D. and Vassilvitskii, S.
\newblock K-means++: The advantages of careful seeding.
\newblock In \emph{Soda}, volume~7, pp.\  1027--1035, 2007.

\bibitem[Barachant et~al.(2011)Barachant, Bonnet, Congedo, and
  Jutten]{barachant2011multiclass}
Barachant, A., Bonnet, S., Congedo, M., and Jutten, C.
\newblock Multiclass brain--computer interface classification by {R}iemannian
  geometry.
\newblock \emph{IEEE Transactions on Biomedical Engineering}, 59\penalty0
  (4):\penalty0 920--928, 2011.

\bibitem[Barachant et~al.(2023)Barachant, Barth{\'e}lemy, Gramfort, KING,
  Rodrigues, Dave, Olivetti, Goncharenko, maxdolle, vom Berg, G.Reguig,
  Yamamoto, Artim436, Beasley, Bj{\"a}reholt, Clisson, H{\"o}chenberger,
  jliersch, Sassenhagen, mccorsi, mhurte, stevenmortier, and
  stonebig]{alexandre_barachant_2023_8059038}
Barachant, A., Barth{\'e}lemy, Q., Gramfort, A., KING, J.-R., Rodrigues, P.
  L.~C., Dave, Olivetti, E., Goncharenko, V., maxdolle, vom Berg, G.~W.,
  G.Reguig, Yamamoto, M.~S., Artim436, Beasley, B., Bj{\"a}reholt, E., Clisson,
  P., H{\"o}chenberger, R., jliersch, Sassenhagen, J., mccorsi, mhurte,
  stevenmortier, and stonebig.
\newblock pyriemann/pyriemann: v0.5, June 2023.
\newblock URL \url{https://doi.org/10.5281/zenodo.8059038}.

\bibitem[Bhatia(2015)]{bhatia}
Bhatia, R.
\newblock \emph{Positive Definite Matrices}.
\newblock Princeton University Press, USA, 2015.
\newblock ISBN 0691168253.

\bibitem[Boumal(2023)]{boumal2023introduction}
Boumal, N.
\newblock \emph{An introduction to optimization on smooth manifolds}.
\newblock Cambridge University Press, 2023.

\bibitem[Brooks et~al.(2019)Brooks, Schwander, Barbaresco, Schneider, and
  Cord]{brooks2019riemannian}
Brooks, D., Schwander, O., Barbaresco, F., Schneider, J.-Y., and Cord, M.
\newblock Riemannian batch normalization for {SPD} neural networks.
\newblock \emph{Advances in Neural Information Processing Systems}, 32, 2019.

\bibitem[Collas et~al.(2021)Collas, Bouchard, Breloy, Ginolhac, Ren, and
  Ovarlez]{9627641}
Collas, A., Bouchard, F., Breloy, A., Ginolhac, G., Ren, C., and Ovarlez, J.-P.
\newblock Probabilistic pca from heteroscedastic signals: Geometric framework
  and application to clustering.
\newblock \emph{IEEE Transactions on Signal Processing}, 69:\penalty0
  6546--6560, 2021.
\newblock \doi{10.1109/TSP.2021.3130997}.

\bibitem[Couillet \& Liao(2022)Couillet and Liao]{couillet2022random}
Couillet, R. and Liao, Z.
\newblock \emph{Random matrix methods for machine learning}.
\newblock Cambridge University Press, 2022.

\bibitem[Couillet et~al.(2019)Couillet, Tiomoko, Zozor, and
  Moisan]{couillet2019random}
Couillet, R., Tiomoko, M., Zozor, S., and Moisan, E.
\newblock Random matrix-improved estimation of covariance matrix distances.
\newblock \emph{Journal of Multivariate Analysis}, 174:\penalty0 104531, 2019.

\bibitem[El~Karoui(2008)]{KAR08}
El~Karoui, N.
\newblock {Spectrum estimation for large dimensional covariance matrices using
  random matrix theory}.
\newblock \emph{Annals of Statistics}, 36\penalty0 (6):\penalty0 2757--2790,
  December 2008.

\bibitem[Fletcher \& Joshi(2004)Fletcher and Joshi]{fletcher2004principal}
Fletcher, P.~T. and Joshi, S.
\newblock Principal geodesic analysis on symmetric spaces: Statistics of
  diffusion tensors.
\newblock In \emph{International Workshop on Mathematical Methods in Medical
  and Biomedical Image Analysis}, pp.\  87--98. Springer, 2004.

\bibitem[Han et~al.(2021)Han, Mishra, Jawanpuria, and
  Gao]{NEURIPS2021_4b04b0dc}
Han, A., Mishra, B., Jawanpuria, P.~K., and Gao, J.
\newblock On riemannian optimization over positive definite matrices with the
  bures-wasserstein geometry.
\newblock In Ranzato, M., Beygelzimer, A., Dauphin, Y., Liang, P., and Vaughan,
  J.~W. (eds.), \emph{Advances in Neural Information Processing Systems},
  volume~34, pp.\  8940--8953. Curran Associates, Inc., 2021.
\newblock URL
  \url{https://proceedings.neurips.cc/paper_files/paper/2021/file/4b04b0dcd2ade339a3d7ce13252a29d4-Paper.pdf}.

\bibitem[Harandi et~al.(2017)Harandi, Salzmann, and
  Hartley]{pmlr-v70-harandi17a}
Harandi, M., Salzmann, M., and Hartley, R.
\newblock Joint dimensionality reduction and metric learning: A geometric take.
\newblock In Precup, D. and Teh, Y.~W. (eds.), \emph{Proceedings of the 34th
  International Conference on Machine Learning}, volume~70 of \emph{Proceedings
  of Machine Learning Research}, pp.\  1404--1413. PMLR, 06--11 Aug 2017.
\newblock URL \url{https://proceedings.mlr.press/v70/harandi17a.html}.

\bibitem[Huang \& Van~Gool(2017)Huang and Van~Gool]{huang2017riemannian}
Huang, Z. and Van~Gool, L.
\newblock A riemannian network for spd matrix learning.
\newblock In \emph{Proceedings of the AAAI conference on artificial
  intelligence}, volume~31, 2017.

\bibitem[Jeuris et~al.(2012)Jeuris, Vandebril, and
  Vandereycken]{jeuris2012survey}
Jeuris, B., Vandebril, R., and Vandereycken, B.
\newblock A survey and comparison of contemporary algorithms for computing the
  matrix geometric mean.
\newblock \emph{Electronic Transactions on Numerical Analysis}, 39:\penalty0
  379--402, 2012.

\bibitem[Kammoun et~al.(2018)Kammoun, Couillet, Pascal, and
  Alouini]{Kammoun2017}
Kammoun, A., Couillet, R., Pascal, F., and Alouini, M.-S.
\newblock Optimal design of the adaptive normalized matched filter detector
  using regularized tyler estimators.
\newblock \emph{IEEE Transactions on Aerospace and Electronic Systems},
  54\penalty0 (2):\penalty0 755--769, 2018.
\newblock \doi{10.1109/TAES.2017.2766538}.

\bibitem[Kobler et~al.(2022)Kobler, ichiro Hirayama, Zhao, and
  Kawanabe]{kobler2022spd}
Kobler, R.~J., ichiro Hirayama, J., Zhao, Q., and Kawanabe, M.
\newblock Spd domain-specific batch normalization to crack interpretable
  unsupervised domain adaptation in eeg.
\newblock In \emph{Neurips}, 2022.

\bibitem[Ledoit \& Wolf(2004)Ledoit and Wolf]{ledoit2004well}
Ledoit, O. and Wolf, M.
\newblock A well-conditioned estimator for large-dimensional covariance
  matrices.
\newblock \emph{Journal of multivariate analysis}, 88\penalty0 (2):\penalty0
  365--411, 2004.

\bibitem[Ledoit \& Wolf(2015)Ledoit and Wolf]{ledoit2015spectrum}
Ledoit, O. and Wolf, M.
\newblock Spectrum estimation: A unified framework for covariance matrix
  estimation and {PCA} in large dimensions.
\newblock \emph{Journal of Multivariate Analysis}, 139:\penalty0 360--384,
  2015.

\bibitem[Ledoit \& Wolf(2018)Ledoit and Wolf]{ledoit2018optimal}
Ledoit, O. and Wolf, M.
\newblock Optimal estimation of a large-dimensional covariance matrix under
  {S}tein’s loss.
\newblock \emph{Bernoulli}, 24\penalty0 (4B):\penalty0 3791--3832, 2018.

\bibitem[Ledoit \& Wolf(2020)Ledoit and Wolf]{ledoit2020analytical}
Ledoit, O. and Wolf, M.
\newblock Analytical nonlinear shrinkage of large-dimensional covariance
  matrices.
\newblock \emph{The Annals of Statistics}, 48\penalty0 (5):\penalty0 3043 --
  3065, 2020.

\bibitem[Lou et~al.(2020)Lou, Katsman, Jiang, Belongie, Lim, and
  De~Sa]{lou2020}
Lou, A., Katsman, I., Jiang, Q., Belongie, S., Lim, S.-N., and De~Sa, C.
\newblock Differentiating through the fr\'{e}chet mean.
\newblock In \emph{Proceedings of the 37th International Conference on Machine
  Learning}, ICML'20. JMLR.org, 2020.

\bibitem[Marchenko \& Pastur(1967)Marchenko and
  Pastur]{marchenko1967distribution}
Marchenko, V.~A. and Pastur, L.~A.
\newblock Distribution of eigenvalues for some sets of random matrices.
\newblock \emph{Matematicheskii Sbornik}, 114\penalty0 (4):\penalty0 507--536,
  1967.

\bibitem[Ollila \& Tyler(2014)Ollila and Tyler]{ollila2014regularized}
Ollila, E. and Tyler, D.~E.
\newblock Regularized $m$-estimators of scatter matrix.
\newblock \emph{IEEE Transactions on Signal Processing}, 62\penalty0
  (22):\penalty0 6059--6070, 2014.

\bibitem[Pascal et~al.(2014)Pascal, Chitour, and Quek]{pascal2014regularized}
Pascal, F., Chitour, Y., and Quek, Y.
\newblock Generalized robust shrinkage estimator and its application to stap
  detection problem.
\newblock \emph{IEEE Transactions on Signal Processing}, 62\penalty0
  (21):\penalty0 5640--5651, 2014.
\newblock \doi{10.1109/TSP.2014.2355779}.

\bibitem[Pedregosa et~al.(2018)Pedregosa, Varoquaux, Gramfort, Michel, Thirion,
  Grisel, Blondel, Müller, Nothman, Louppe, Prettenhofer, Weiss, Dubourg,
  Vanderplas, Passos, Cournapeau, Brucher, Perrot, and Édouard
  Duchesnay]{pedregosa2018scikitlearn}
Pedregosa, F., Varoquaux, G., Gramfort, A., Michel, V., Thirion, B., Grisel,
  O., Blondel, M., Müller, A., Nothman, J., Louppe, G., Prettenhofer, P.,
  Weiss, R., Dubourg, V., Vanderplas, J., Passos, A., Cournapeau, D., Brucher,
  M., Perrot, M., and Édouard Duchesnay.
\newblock Scikit-learn: Machine learning in python, 2018.

\bibitem[Pereira et~al.(2023)Pereira, Mestre, and
  Gregoratti]{pereira2023consistent}
Pereira, R., Mestre, X., and Gregoratti, D.
\newblock Consistent estimators of a new class of covariance matrix distances
  in the large dimensional regime.
\newblock In \emph{ICASSP 2023-2023 IEEE International Conference on Acoustics,
  Speech and Signal Processing (ICASSP)}, pp.\  1--5. IEEE, 2023.

\bibitem[Reimherr et~al.(2021)Reimherr, Bharath, and Soto]{reimherr2021}
Reimherr, M.~L., Bharath, K., and Soto, C.
\newblock Differential privacy over riemannian manifolds.
\newblock In \emph{Neural Information Processing Systems}, 2021.

\bibitem[Ru{\ss}wurm et~al.(2020)Ru{\ss}wurm, Pelletier, Zollner, Lef{\`e}vre,
  and K{\"o}rner]{breizhcrops2020}
Ru{\ss}wurm, M., Pelletier, C., Zollner, M., Lef{\`e}vre, S., and K{\"o}rner,
  M.
\newblock Breizhcrops: A time series dataset for crop type mapping.
\newblock \emph{International Archives of the Photogrammetry, Remote Sensing
  and Spatial Information Sciences ISPRS (2020)}, 2020.

\bibitem[Silverstein \& Bai(1995)Silverstein and Bai]{silverstein1995empirical}
Silverstein, J.~W. and Bai, Z.
\newblock On the empirical distribution of eigenvalues of a class of large
  dimensional random matrices.
\newblock \emph{Journal of Multivariate analysis}, 54\penalty0 (2):\penalty0
  175--192, 1995.

\bibitem[Skovgaard(1984)]{skovgaard1984riemannian}
Skovgaard, L.~T.
\newblock A {R}iemannian geometry of the multivariate normal model.
\newblock \emph{Scandinavian journal of statistics}, pp.\  211--223, 1984.

\bibitem[Tiomoko et~al.(2019)Tiomoko, Couillet, Bouchard, and
  Ginolhac]{tiomoko2019random}
Tiomoko, M., Couillet, R., Bouchard, F., and Ginolhac, G.
\newblock Random matrix improved covariance estimation for a large class of
  metrics.
\newblock In \emph{International Conference on Machine Learning}, pp.\
  6254--6263. PMLR, 2019.

\bibitem[Tuzel et~al.(2008)Tuzel, Porikli, and Meer]{tuzel2008pedestrian}
Tuzel, O., Porikli, F., and Meer, P.
\newblock Pedestrian detection via classification on {R}iemannian manifolds.
\newblock \emph{IEEE transactions on pattern analysis and machine
  intelligence}, 30\penalty0 (10):\penalty0 1713--1727, 2008.

\bibitem[Utpala et~al.(2023)Utpala, Vepakomma, and Miolane]{utpala2023}
Utpala, S., Vepakomma, P., and Miolane, N.
\newblock Differentially private fr\'echet mean on the manifold of symmetric
  positive definite ({SPD}) matrices with log-euclidean metric.
\newblock \emph{Transactions on Machine Learning Research}, 2023.
\newblock ISSN 2835-8856.

\bibitem[Wishart(1928)]{wishart1928generalised}
Wishart, J.
\newblock The generalised product moment distribution in samples from a normal
  multivariate population.
\newblock \emph{Biometrika}, pp.\  32--52, 1928.

\bibitem[Zadeh et~al.(2016)Zadeh, Hosseini, and Sra]{zadeh2016}
Zadeh, P., Hosseini, R., and Sra, S.
\newblock Geometric mean metric learning.
\newblock In Balcan, M.~F. and Weinberger, K.~Q. (eds.), \emph{Proceedings of
  The 33rd International Conference on Machine Learning}, volume~48 of
  \emph{Proceedings of Machine Learning Research}, pp.\  2464--2471, New York,
  New York, USA, 20--22 Jun 2016. PMLR.
\newblock URL \url{https://proceedings.mlr.press/v48/zadeh16.html}.

\end{thebibliography}

\newpage
\appendix
\onecolumn

\section{Proofs of Propositions~\ref{prop:Rgrad_eigsfun} and~\ref{prop:RMT_Fisher_SCM_deterministic_grad}}
\label{app:proofs}

\subsection{Proof of Proposition~\ref{prop:Rgrad_eigsfun}}

Let $f(\point) = g(\eigsMat)$, where we have the eigenvalue decomposition $\point^{\nicefrac{-1}2}\SCM\point^{\nicefrac{-1}2}=\MAT{U}\eigsMat\MAT{U}^T$.
By definition, we have
$$
    \diff f(\point) = \tr(\point^{-1}\grad f(\point)\point^{-1}\diff\point)
    = \diff g(\eigsMat) = \tr(\eigsMat^{-1}\grad g(\eigsMat)\eigsMat^{-1}\diff\eigsMat).
$$
$\diff\eigsMat$ corresponds to the differential of the eigenvalues of $\point^{\nicefrac{-1}2}\SCM\point^{\nicefrac{-1}2}$, which is known to be equal to $\diff\eigsMat=\diag(\MAT{U}^T\diff(\point^{\nicefrac{-1}2}\SCM\point^{\nicefrac{-1}2})\MAT{U})$.
For any matrix $\MAT{N}$ and diagonal matrix $\MAT{\Delta}$, we have $\tr(\MAT{\Delta}\diag(\MAT{N})) = \tr(\MAT{\Delta}\MAT{N})$.
Hence, since $\eigsMat^{-1}\grad g(\eigsMat)\eigsMat^{-1}$ is diagonal, we obtain
$$
     \tr(\eigsMat^{-1}\grad g(\eigsMat)\eigsMat^{-1}\diag(\MAT{U}^T\diff(\point^{\nicefrac{-1}2}\SCM\point^{\nicefrac{-1}2})\MAT{U}))
     =
     \tr(\eigsMat^{-1}\grad g(\eigsMat)\eigsMat^{-1}\MAT{U}^T\diff(\point^{\nicefrac{-1}2}\SCM\point^{\nicefrac{-1}2})\MAT{U}).
$$
We further have
$\diff(\point^{\nicefrac{-1}2}\SCM\point^{\nicefrac{-1}2})=\diff(\point^{\nicefrac{-1}2})\SCM\point^{\nicefrac{-1}2} + \point^{\nicefrac{-1}2}\SCM\diff(\point^{\nicefrac{-1}2})$,
where $\diff(\point^{\nicefrac{-1}2})$ is the unique symmetric solution to the equation
$\diff(\point^{\nicefrac{-1}2})\point^{\nicefrac{-1}2} + \point^{\nicefrac{-1}2}\diff(\point^{\nicefrac{-1}2}) = \diff(\point^{-1})=-\point^{-1}\diff\point\point^{-1}$.
Let $\MAT{M}=\point^{\nicefrac{-1}2}\SCM\point^{\nicefrac{-1}2}$ and $\MAT{D} = \eigsMat^{-1}\grad g(\eigsMat)$.
Since $\MAT{U}\MAT{U}^T=\eye$, we have $\MAT{U}\eigsMat^{-1}\grad g(\eigsMat)\eigsMat^{-1}\MAT{U}^T=\MAT{M}^{-1}\MAT{U}\MAT{D}\MAT{U}^T=\MAT{U}\MAT{D}\MAT{U}^T\MAT{M}^{-1}$.
From there, we obtain
$$
    \diff g(\eigsMat) =
    \tr(\MAT{U}\MAT{D}\MAT{U}^T(\diff(\point^{\nicefrac{-1}2})\point^{\nicefrac12} + \point^{\nicefrac12}\diff(\point^{\nicefrac{-1}2}))).
$$
Leveraging $\diff(\point^{\nicefrac{-1}2})\point^{\nicefrac{-1}2} + \point^{\nicefrac{-1}2}\diff(\point^{\nicefrac{-1}2}) = -\point^{-1}\diff\point\point^{-1}$, one gets
$$
    \diff g(\eigsMat) = -\tr(\point^{\nicefrac{-1}2}\MAT{U}\MAT{D}\MAT{U}^T\point^{\nicefrac{-1}2}\diff\point) =
    \diff f(\point) = \tr(\point^{-1}\grad f(\point)\point^{-1}\diff\point).
$$
The result is finally obtained by identification.

\subsection{Proof of Proposition~\ref{prop:RMT_Fisher_SCM_deterministic_grad}}

To get the gradient of $g$, the directional derivative $\diff g(\eigsMat)$ at $\eigsMat$ is computed.
Let the eigenvalue decomposition $\eigsMat - \frac{\sqrt{\VEC{\lambda}}\sqrt{\VEC{\lambda}}^T}{\nsamples}=\MAT{V}\diag(\VEC{\zeta})\MAT{V}^T$, where $\VEC{\lambda}=\diag(\eigsMat)$.
The differential $\diff\VEC{\zeta}$ of the eigenvalues $\VEC{\zeta}$ is $\diff\VEC{\zeta}=\diag(\MAT{V}^T\diff(\eigsMat - \frac{\sqrt{\VEC{\lambda}}\sqrt{\VEC{\lambda}}^T}{\nsamples})\MAT{V})$.
Differentiating each term of $g$ yields
\begin{equation*}
    \diff g(\MAT{\Lambda}) = \frac{1}{2p} \diff (\tr(\log^2(\MAT{\Lambda}))) + \frac{1}{p} \diff (\log|\MAT{\Lambda}|) - (\diff \VEC{\lambda}-\diff \VEC{\zeta})^T\left( \frac{1}{p} \MAT{Q}\onevec +\frac{1-c}{c} \VEC{q} \right) - (\VEC{\lambda}-\VEC{\zeta})^T\left( \frac{1}{p} \diff\MAT{Q}\onevec
    +\frac{1-c}{c} \diff\VEC{q} \right).
\end{equation*}
By leveraging classical results, we obtain
\begin{equation*}
    \frac{1}{2p} \diff (\tr(\log^2(\MAT{\Lambda}))) + \frac{1}{p} \diff (\log|\MAT{\Lambda}|)
    =
    \frac1p\tr([\log(\eigsMat)+\eye]\eigsMat^{-1}\diff\eigsMat).
\end{equation*}
In the following, $\div$ corresponds to the element-wise division, $\odot$ denotes the Hadamard (element-wise) product, and $\cdot^{\odot\cdot}$ corresponds to the element-wise power function.
From~\eqref{eq:RMT_Fisher_SCM_deterministic}, $\VEC{q}=\diag(\log(\eigsMat)\eigsMat^{-1})=\frac{\log\VEC{\lambda}}{\VEC{\lambda}}$, and we obtain $\diff\VEC{q}=\diag((\eye-\log\eigsMat)\eigsMat^{-2}\diff\eigsMat)=\frac{\onevec-\log\VEC{\lambda}}{\VEC{\lambda}^{\odot 2}}\odot\diff\VEC{\lambda}$.
One can also rewrite $\MAT{Q}$ as
\begin{equation*}
    \MAT{Q} = \frac{[(\VEC{\lambda}\odot\log\VEC{\lambda})\cdot\onevec^T - \VEC{\lambda}\cdot\log\VEC{\lambda}^T] - [\VEC{\lambda}\cdot\onevec^T - \onevec\cdot\VEC{\lambda}^T] + \eye}{[\VEC{\lambda}\cdot\onevec^T - \onevec\cdot\VEC{\lambda}^T]^{\odot 2} + 2\eigsMat}.
\end{equation*}
Differentiating this yields
\begin{multline*}
    \diff\MAT{Q} =
    \frac{
        \diff\eigsMat[\log\VEC{\lambda}\cdot\onevec^T - \onevec\cdot\log\VEC{\lambda}^T] + [\VEC{\lambda}\cdot\onevec^T-\onevec\cdot\VEC{\lambda}^T]\odot(\onevec\cdot\frac{\onevec}{\VEC{\lambda}}^T)\diff\eigsMat
    }
    {
        [\VEC{\lambda}\cdot\onevec^T-\onevec\cdot\VEC{\lambda}^T]^{\odot 2} + 2\eigsMat
    }
    \\
    -\frac{
        (2\diff\eigsMat(\onevec\cdot\onevec^T) + [\eye - 2(\onevec\cdot\onevec^T)]\diff\eigsMat)
        \odot
        ([(\VEC{\lambda}\odot\log\VEC{\lambda})\cdot\onevec^T - \VEC{\lambda}\cdot\log\VEC{\lambda}^T] - [\VEC{\lambda}\cdot\onevec^T - \onevec\cdot\VEC{\lambda}^T] + \eye)
    }
    {
        [\VEC{\lambda}\cdot\onevec^T-\onevec\cdot\VEC{\lambda}^T]^{\odot 3} + 2\eigsMat^2
    },
\end{multline*}
where we use $(\diff\VEC{\lambda}\odot\VEC{a})\cdot\VEC{b}^T=\diff\eigsMat(\VEC{a}\cdot\VEC{b}^T)$ and $\VEC{a}\cdot(\diff\VEC{\lambda}\odot\VEC{b})^T=(\VEC{a}\cdot\VEC{b}^T)\diff\eigsMat$.
Notice that in the equation above, when the diagonal part of the numerator is equal to zero, then the diagonal part of the denominator can be replaced with anything different from zero.
We usually choose $\eye$.
From there, calculations allow to obtain $\diff\MAT{Q}=\diff\eigsMat\MAT{B}+\MAT{C}\diff\eigsMat$ with $\MAT{B}$ and $\MAT{C}$ defined in Proposition \ref{prop:RMT_Fisher_SCM_deterministic_grad}.

Further calculations yield
\begin{equation*}
    - (\diff \VEC{\lambda}-\diff \VEC{\zeta})^T\left( \frac{1}{p} \MAT{Q}\onevec +\frac{1-c}{c} \VEC{q} \right) = -\tr(\MAT{\Delta}\diff\eigsMat) - \tr(\diag(\MAT{A}\MAT{V}\MAT{\Delta}\MAT{A}^T)\diff\eigsMat),
\end{equation*}
where $\MAT{A}$ and $\MAT{\Delta}$ are defined in proposition \ref{prop:RMT_Fisher_SCM_deterministic_grad}.
We also have
$$
    - \frac{1}{p}(\VEC{\lambda}-\VEC{\zeta})^T\diff\MAT{Q}\onevec
    = -\frac1p\tr(\diag(\MAT{B}\onevec(\VEC{\lambda}-\VEC{\zeta})^T + \onevec(\VEC{\lambda}-\VEC{\zeta})^T\MAT{C})\diff\eigsMat),
$$
and
$$
    -\frac{1-c}{c}(\VEC{\lambda}-\VEC{\zeta})^T\diff\VEC{q}
    = -\frac{1-c}{c}\tr(\eigsMat^{-2}(\eye-\log(\eigsMat))(\eigsMat-\diag(\zeta))\diff\eigsMat).
$$
The result is obtained by combining all above equation and identification with $\tr(\eigsMat^{-2}\grad g(\eigsMat)\diff\eigsMat) = \diff g(\MAT{\Lambda})$.

\section{Simulations for covariance estimation of Section~\ref{sec:cov}}
\label{app:simu_RMTCov}
The experimental setting is as follows: some random covariance $\Cov=\MAT{U}\MAT{\Delta}\MAT{U}^T\in\SPDman$ ($\nfeatures=64$) is generated, where $\MAT{U}$ is uniformly drawn on $\mathcal{O}_{\nfeatures}$ (orthogonal group), and $\MAT{\Delta}$ is randomly drawn on $\DPDman$.
Maximal and minimal diagonal entries of $\MAT{\Delta}$ are set to $\sqrt{a}$ and $\nicefrac{1}{\sqrt{a}}$, where $a=100$ is the condition number.
Remaining non-zero elements are uniformly drawn in-between.
From there, matrices $\dataMat\in\realSpace^{\nfeatures\times\nsamples}$ are simulated.
Each column vector of $\dataMat$ is independently drawn from $\mathcal{N}(\VEC{0},\Cov)$.
The effect of the number of samples $\nsamples$ is studied.
We perform $1000$ Monte Carlo simulations.

To estimate $\Cov$ from $\dataMat$, we consider the following methods:
(\emph{i}) the SCM estimator $\SCM$;
(\emph{ii}) the linear Ledoit-Wolf estimator $\linearLW$~\cite{ledoit2004well}
(\emph{ii}) the non-linear Ledoit-Wolf estimator $\nonlinearLW$~\cite{ledoit2020analytical}
and (\emph{iii}) our RMT distance based method $\CovRMTdist$ from Algorithm~\ref{algo:RMTCov}.
To measure performance, we evaluate the squared Fisher distance~\eqref{eq:Fisher_dist} between $\Cov$ and the estimators.

\begin{figure}[hb!]
    \centering
    \input{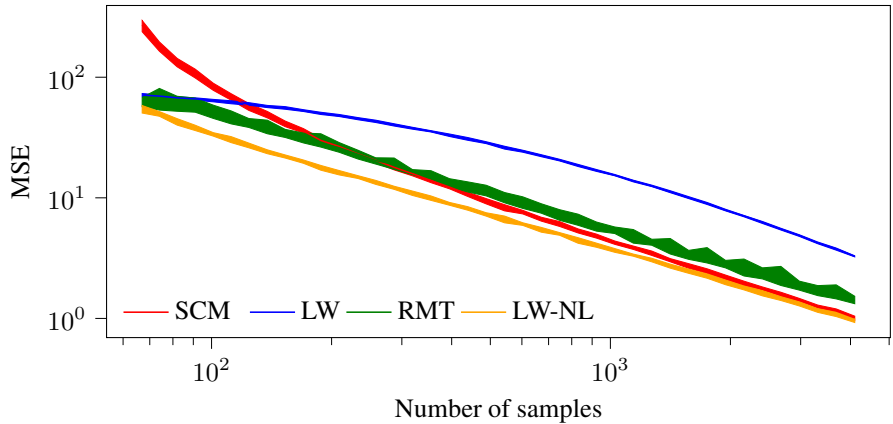}
    \caption{MSE of the estimated covariance. Parameters are $p=64$, $\ell_{\mathrm{max}}=100$, $\epsilon=10^{-6}$, $\alpha=10$. Plot done over 1000 trials. The line corresponds to the median and the filled area corresponds to the $5$-th and $95$-th quantiles over the trials.}
    \label{fig:cov_simu}
\end{figure}

Results are given in Figure~\ref{fig:cov_simu}.
We observe that the best performance is obtained by $\nonlinearLW$.
Our estimator $\CovRMTdist$ improves upon $\SCM$ and $\linearLW$ at low sample support.
From these results, it does not appear appealing.
It is also computationally significantly more expensive than other estimators, which are analytically known.
Thus, exploiting~\eqref{eq:RMT_Fisher_SCM_deterministic} might generally not be suited for covariance estimation.
To conclude on a positive note, notice that, while conducting our simulations, we encountered some rare cases at low sample support where $\SCM$, $\linearLW$ and $\nonlinearLW$ behave poorly (especially $\nonlinearLW$), while $\CovRMTdist$ performed well.
We believe that this occurs when the SCM does not provide good eigenvectors.

\end{document}